\documentclass[a4paper,UKenglish]{lipics-v2016}

\usepackage{hyperref}
\usepackage{enumerate}
\usepackage[color]{xypic}
\input xy
\xyoption{all}
\usepackage{comment}
\usepackage{mathtools}
\usepackage{fancybox}
\usepackage{nicefrac}
\usepackage{xspace}

\usepackage{tikz}
\usetikzlibrary{circuits.ee.IEC}
\usetikzlibrary{shapes,shapes.geometric,shapes.misc}
\newlength\stateheight
\newlength\minimumstatewidth
\tikzset{width/.initial=\minimummorphismwidth}
\tikzset{colour/.initial=white}

\newif\ifblack\pgfkeys{/tikz/black/.is if=black}
\newif\ifwedge\pgfkeys{/tikz/wedge/.is if=wedge}
\newif\ifvflip\pgfkeys{/tikz/vflip/.is if=vflip}
\newif\ifhflip\pgfkeys{/tikz/hflip/.is if=hflip}
\newif\ifhvflip\pgfkeys{/tikz/hvflip/.is if=hvflip}

\pgfdeclarelayer{foreground}
\pgfdeclarelayer{background}
\pgfsetlayers{background,main,foreground}

\def\thickness{0.4pt}

\makeatletter
\pgfkeys{%
  /tikz/on layer/.code={
    \pgfonlayer{#1}\begingroup
    \aftergroup\endpgfonlayer
    \aftergroup\endgroup
  },
  /tikz/node on layer/.code={
    \gdef\node@@on@layer{%
      \setbox\tikz@tempbox=\hbox\bgroup\pgfonlayer{#1}\unhbox\tikz@tempbox\endpgfonlayer\pgfsetlinewidth{\thickness}\egroup}
    \aftergroup\node@on@layer
  },
  /tikz/end node on layer/.code={
    \endpgfonlayer\endgroup\endgroup
  }
}
\def\node@on@layer{\aftergroup\node@@on@layer}
\makeatother

\makeatletter
\pgfdeclareshape{state}
{
    \savedanchor\centerpoint
    {
        \pgf@x=0pt
        \pgf@y=0pt
    }
    \anchor{center}{\centerpoint}
    \anchorborder{\centerpoint}
    \saveddimen\overallwidth
    {
        \pgf@x=3\wd\pgfnodeparttextbox
        \ifdim\pgf@x<\minimumstatewidth
            \pgf@x=\minimumstatewidth
        \fi
    }
    \savedanchor{\upperrightcorner}
    {
        \pgf@x=.5\wd\pgfnodeparttextbox
        \pgf@y=.5\ht\pgfnodeparttextbox
        \advance\pgf@y by -.5\dp\pgfnodeparttextbox
    }
    \anchor{A}
    {
        \pgf@x=-\overallwidth
        \divide\pgf@x by 4
        \pgf@y=0pt
    }
    \anchor{B}
    {
        \pgf@x=\overallwidth
        \divide\pgf@x by 4
        \pgf@y=0pt
    }
    \anchor{text}
    {
        \upperrightcorner
        \pgf@x=-\pgf@x
        \ifhflip
            \pgf@y=-\pgf@y
            \advance\pgf@y by 0.4\stateheight
        \else
            \pgf@y=-\pgf@y
            \advance\pgf@y by -0.4\stateheight
        \fi
    }
    \backgroundpath
    {
       \begin{pgfonlayer}{foreground}
        \pgfsetstrokecolor{black}
        \pgfsetlinewidth{\thickness}
        \pgfpathmoveto{\pgfpoint{-0.5*\overallwidth}{0}}
        \pgfpathlineto{\pgfpoint{0.5*\overallwidth}{0}}
        \ifhflip
            \pgfpathlineto{\pgfpoint{0}{\stateheight}}
        \else
            \pgfpathlineto{\pgfpoint{0}{-\stateheight}}
        \fi
        \pgfpathclose
        \ifblack
            \pgfsetfillcolor{black!50}
            \pgfusepath{fill,stroke}
        \else
            \pgfusepath{stroke}
        \fi
       \end{pgfonlayer}
    }
}

\pgfdeclareshape{my ground}
{
  \inheritsavedanchors[from=rectangle ee]
  \inheritanchor[from=rectangle ee]{center}
  \inheritanchor[from=rectangle ee]{north}
  \inheritanchor[from=rectangle ee]{south}
  \inheritanchor[from=rectangle ee]{east}
  \inheritanchor[from=rectangle ee]{west}
  \inheritanchor[from=rectangle ee]{north east}
  \inheritanchor[from=rectangle ee]{north west}
  \inheritanchor[from=rectangle ee]{south east}
  \inheritanchor[from=rectangle ee]{south west}
  \inheritanchor[from=rectangle ee]{input}
  \inheritanchor[from=rectangle ee]{output}
  \anchorborder{\csname pgf@anchor@my ground@input\endcsname}

  \backgroundpath{
    \pgf@process{\pgfpointadd{\southwest}{\pgfpoint{\pgfkeysvalueof{/pgf/outer xsep}}{\pgfkeysvalueof{/pgf/outer ysep}}}}
    \pgf@xa=\pgf@x \pgf@ya=\pgf@y
    \pgf@process{\pgfpointadd{\northeast}{\pgfpointscale{-1}{\pgfpoint{\pgfkeysvalueof{/pgf/outer xsep}}{\pgfkeysvalueof{/pgf/outer ysep}}}}}
    \pgf@xb=\pgf@x \pgf@yb=\pgf@y
    \pgfpathmoveto{\pgfqpoint{\pgf@xa}{\pgf@ya}}
    \pgfpathlineto{\pgfqpoint{\pgf@xa}{\pgf@yb}}
    \pgfmathsetlength\pgf@xa{.5\pgf@xa+.5\pgf@xb}
    \pgfmathsetlength\pgf@yc{.16666\pgf@yb-.16666\pgf@ya}
    \advance\pgf@ya by\pgf@yc
    \advance\pgf@yb by-\pgf@yc
    \pgfpathmoveto{\pgfqpoint{\pgf@xa}{\pgf@ya}}
    \pgfpathlineto{\pgfqpoint{\pgf@xa}{\pgf@yb}}
    \advance\pgf@ya by\pgf@yc
    \advance\pgf@yb by-\pgf@yc
    \pgfpathmoveto{\pgfqpoint{\pgf@xb}{\pgf@ya}}
    \pgfpathlineto{\pgfqpoint{\pgf@xb}{\pgf@yb}}
  }
}
\makeatother

\tikzset{inline text/.style =
  {text height=1.2ex, text depth=0.25ex,yshift=0.5mm}}
\tikzset{arrow box/.style =
  {rectangle,inline text,fill=white,draw,
    minimum height=5mm,yshift=-0.5mm,minimum width=5mm}}
\tikzset{dot/.style =
  {inner sep=0mm,minimum width=1mm,minimum height=1mm,
    draw,shape=circle}}
\tikzset{white dot/.style = {dot,fill=white,text depth=-0.2mm}}
\tikzset{scalar/.style = {diamond,draw,inner sep=1pt}}

\tikzset{copier/.style = {dot,fill,text depth=-0.2mm}}
\tikzset{discarder/.style = {my ground,draw,inner sep=0pt,
    minimum width=4.2pt,minimum height=11.2pt,anchor=input,rotate=90}}

\tikzset{uniform/.style = {my ground,draw,inner sep=0pt,
    minimum width=4.2pt,minimum height=11.2pt,anchor=input,rotate=270}}

\tikzset{xshiftu/.style = {shift = {(#1, 0)}}}
\tikzset{yshiftu/.style = {shift = {(0, #1)}}}

\renewcommand{\arraystretch}{1.3}
\newcommand{\QEDbox}{\qedsymbol} 

\usepackage[layout=margin,draft]{fixme}
\FXRegisterAuthor{fz}{afz}{Fab}
\FXRegisterAuthor{bj}{abj}{Bar}

\newif\ifignore 
\ignorefalse
\newcommand{\auxproof}[1]{
\ifignore\mbox{}\newline
\textbf{PROOF:} \dotfill\newline
{\it #1}\mbox{}\newline
\textbf{ENDPROOF}\dotfill
\fi}

\def \df {\mathrel{\ensuremath{:=}}} 

\def \: {\colon} 
\def \after {\mathrel{\circ}}
\newcommand{\idmap}[1][]{\ensuremath{\mathrm{id}_{#1}}}

\newsavebox\sbpto
\savebox\sbpto{\begin{tikzpicture}[baseline=-2.5pt]
            \filldraw[fill=white,draw=white] circle (1.4pt);
            \filldraw[fill=white,draw=black,line width=0.2pt] circle
(1.2pt);
                \end{tikzpicture}}
\newcommand\pto{\mathrel{\ooalign{$\to$\cr
            \hfil\!$\usebox\sbpto$\hfil\cr}}}
            
\newcommand\kto[2]{#1 \pto #2}

\newcommand{\Wcirc}{\lower5pt\hbox{$\includegraphics[height=5pt]{graffles/Wcircle.pdf}$}}

\newsavebox\sbpafterd
\savebox\sbpafterd{\begin{tikzpicture}[baseline=-2.5pt]
            \filldraw[fill=white,draw=white] circle (2.1pt);
            \filldraw[fill=white,draw=black, line width=0.2pt] circle (1.8pt);
            \filldraw[fill=black, line width=0] circle (0.2pt);
                                \end{tikzpicture}}

\newsavebox\sbpaftert
\savebox\sbpaftert{\begin{tikzpicture}[baseline=-2.5pt]
            \filldraw[fill=white,draw=white] circle (2.1pt);
            \filldraw[fill=white,draw=black, line width=0.2pt] circle (1.8pt);
            \filldraw[fill=black, line width=0] circle (0.2pt);
                                \end{tikzpicture}}
\newsavebox\sbpafters
\savebox\sbpafters{\begin{tikzpicture}[baseline=-1.75pt]
            \filldraw[fill=white,draw=white] circle (1.7pt);
            \filldraw[fill=white,draw=black,line width=0.15pt] circle (1.4pt);
            \filldraw[fill=black, line width=0] circle (0.15pt);
                                \end{tikzpicture}}

\newsavebox\sbpafterss
\savebox\sbpafterss{\begin{tikzpicture}[baseline=-1.25pt]
            \filldraw[fill=white,draw=white] circle (1.4pt);
            \filldraw[fill=white,draw=black,line width=0.1pt] circle (1.1pt);
            \filldraw[fill=black, line width=0] circle (0.15pt);
                                \end{tikzpicture}}

\newcommand{\klafter}{\begingroup
        \mathchoice{\mathbin{\usebox\sbpafterd}}
        {\mathbin{\usebox\sbpaftert}}
        {\mathbin{\usebox\sbpafters}}
        {\mathbin{\usebox\sbpafterss}}\endgroup}

\newcommand{\setin}[3]{\{#1\in#2\;|\;#3\}}

\newcommand\Dst{\mathcal{D}}

\newcommand{\ket}[1]{\ensuremath{|{\kern.1em}#1{\kern.1em}\rangle}}
\newcommand{\bigket}[1]{\ensuremath{\big|{\kern.1em}#1{\kern.1em}\big\rangle}}
\newcommand{\supp}{\mathrm{supp}}

\newcommand{\Kl}[1]{\mathcal{K}{\kern-.4ex}\ell(#1)}
\newcommand{\intd}{{\kern.2em}\mathrm{d}{\kern.03em}}
\DeclareSymbolFont{T1op}{T1}{cmr}{m}{n}
\SetSymbolFont{T1op}{bold}{T1}{cmr}{bx}{n}
\DeclareMathSymbol{\mathguilsinglleft}{\mathopen}{T1op}{'016}
\DeclareMathSymbol{\mathguilsinglright}{\mathclose}{T1op}{'017}
\newcommand{\klin}[1]{\mathguilsinglleft#1\mathguilsinglright}

\newcommand{\one}{\ensuremath{\mathbf{1}}}
\newcommand{\zero}{\ensuremath{\mathbf{0}}}
\newcommand{\indic}[1]{\one_{#1}}

\newcommand{\andthen}{\mathrel{\&}}

\newcommand{\margsign}{\mathsf{M}}
\newcommand{\rotM}{\rotatebox[origin=c]{180}{$\margsign$}}
\newcommand{\weaksign}{\mathsf{\rotM}}
\newcommand{\fW}{\ensuremath{\weaksign_{1}}}
\newcommand{\subfW}{\ensuremath{{\scriptsize\weaksign}_{1}}} 
\newcommand{\sW}{\ensuremath{\weaksign_{2}}}
\newcommand{\subsW}{\ensuremath{{\scriptsize\weaksign}_{2}}} 
\newcommand{\fM}{\ensuremath{\margsign_{1}}}
\newcommand{\sM}{\ensuremath{\margsign_{2}}}

\newcommand\Idnet{\lower3pt\hbox{$\includegraphics[width=20pt]{graffles/id.pdf}$}}
\newcommand\symNet{\lower3pt\hbox{$\includegraphics[width=20pt]{graffles/symmetryalt.pdf}$}}
\newcommand{\ZeronetT}{\lower4pt\hbox{$\includegraphics[width=14pt]{graffles/idzerocircuit.pdf}$}}

\newsavebox\sbground
\savebox\sbground{%
  \begin{tikzpicture}[baseline=0pt]
    \draw (0,-.1ex) to (0,.85ex)
    node[ground IEC,draw,anchor=input,inner sep=0pt,
    minimum width=3.15pt,minimum height=8.4pt,rotate=90] {};
  \end{tikzpicture}%
}
\newcommand{\ground}{\mathord{\usebox\sbground}}

\newsavebox\sbcopier
\savebox\sbcopier{%
  \begin{tikzpicture}[baseline=0pt]
    \node[copier,scale=.7] (a) at (0,3.6pt) {};
    \draw (a) -- +(-90:.16);
    \draw (a) -- +(45:.19);
    \draw (a) -- +(135:.19);
  \end{tikzpicture}}
\newcommand{\copier}{\mathord{\usebox\sbcopier}}

\newcommand{\EfProb}{\textsc{EfProb}\xspace}

\newcommand{\Prob}{\footnotesize \mathrm{Pr}}

\newcommand{\no}[1]{#1^{\scriptscriptstyle \bot}} 

\newenvironment{myproof}[1][Proof]%
   { \begin{trivlist}%
     \item[\hskip \labelsep {\bfseries #1}]%
   }%
   { \end{trivlist}%
   }

\theoremstyle{plain}
\newtheorem{proposition}[theorem]{Proposition}
\newtheorem{rem}{Remark}

\newcommand\cind[3]{#1 \bot #2 \mid #3} 

\title{\fontsize{15}{18} \selectfont The Logical Essentials of Bayesian Reasoning}

\author[1]{Bart Jacobs}
\author[2]{Fabio Zanasi}

\affil[1]{Radboud Universiteit, Nijmegen, The Netherlands}
\affil[2]{University College London, London, United Kingdom}
\titlerunning{Probabilistic Reasoning}
\authorrunning{B. Jacobs and F. Zanasi} 
\Copyright{Bart Jacobs and Fabio Zanasi}
\subjclass{Probabilistic computation, F.~1.2}

\begin{document}

\maketitle

\begin{abstract}
This chapter offers an accessible introduction to the channel-based
approach to Bayesian probability theory. This framework rests on
algebraic and logical foundations, inspired by the methodologies of
programming language semantics. It offers a uniform, structured and
expressive language for describing Bayesian phenomena in terms of
familiar programming concepts, like channel, predicate transformation
and state transformation. The introduction also covers inference in
Bayesian networks, which will be modelled by a suitable calculus of
string diagrams.
\end{abstract}

\section{Introduction}\label{sec:intro}

In traditional imperative programming one interprets a program as a
function that changes states. Intuitively, the notion of `state'
captures the state of affairs in a computer, as given for instance by
the contents of the relevant parts of the computer's memory. More
abstractly, a program is interpreted as a \emph{state transformer}.
An alternative, logical perspective is to interpret a program as a
\emph{predicate transformer}. In that case the program turns one
predicate into a new predicate. This works in opposite direction: the
program turns a predicate on the `post-state' into a predicate on the
`pre-state', for instance via the weakest precondition computation.
As discovered in the early days of programming semantics, basic
relations exists between state transformation and predicate
transformation, see for instance~\cite{DijkstraS90,Dijkstra:1997}
(formulated in Proposition~\ref{prop:validitytransformation} below).

A similar theory of state and predicate transformation has been
developed for probabilistic programming, see~\cite{Kozen81,Kozen85}.
This approach has been generalised and re-formulated in recent years
in categorical terms, typically using so-called Kleisli
categories~\cite{Jacobs17b} and more generally via the notion of
effectus~\cite{ChoJWW15}. Category theory provides a fundamental
language for the semantics of programming languages. This is clear in
approaches based on domain theory. For instance, many constructions
for types in programming languages have categorical counterparts, like
(co)products, exponentials, and initial algebras (and final
coalgebras) --- where these (co)algebras are used for fixed
points. These categorical notions come with universal properties that
guide the design (syntax) and rules of programming languages.

This use of category theory is well-established in functional
programming languages. However, it is less established in
probabilistic programming. The description of some of the basic
notions of probability theory in categorical terms goes back to the
early 1980s (see~\cite{Giry82}) and has seen a steady stream of
activities since --- see
\textit{e.g.}~\cite{JonesP89,JungT98,VinkR99,BartelsSV04,TixKP05,VaraccaW06,Keimel08,KeimelP09,Panangaden09,Sokolova11,Mislove12,Fong12,CulbertsonS14,ScibiorGG15,StatonYHKW16,ScibiorKVSYCOMHG18}). This
perspective is not a goal in itself, but it does offer a structural,
implementation-independent way of thinking which is natural for
systematic programmers.

This paper offers an introduction to this principled perspective on
probability theory, esp.\ for Bayesian probabilistic programming,
based on earlier work of the authors' in this direction, see
\textit{e.g.}~\cite{Jacobs11c,Jacobs13a,JacobsZ16,Jacobs17a,JacobsZ17}. Even
though it is categorically-driven, our exposition does not require any
categorical prerequisite. The reader interested in an explicitly
categorical description of the framework may consult
\cite{JacobsZ16,ChoJ17a}.

The fundamental concept will be called \emph{channel}: all the basics
of Bayesian probability theory (event, belief revision, Bayesian
network, disintegration, \ldots) will be derived from this single
primitive. In analogy with approaches in programming language
semantics, channels are formally definable as arrows of a certain
Kleisli category: depending on the category of choice, the derived
notions instantiate to discrete or to continuous probability theory
--- and even to quantum probability too, although the quantum world is
out of scope here. This setting does not only provides a completely
\emph{uniform} mathematical description for a variety of phenomena,
but also introduces in Bayesian reasoning fundamental principles of
programming theory, such as \emph{compositionality}: channels are
composable in a variety of ways, resulting in a structured and modular
theory. Furthermore, we argue that the channel-based perspective
improves traditional approaches. We will study scenarios in which the
established language for describing probabilistic phenomena lacks in
flexibility, expressiveness and rigour, while the new foundations
disclose the underlying logical structure of phenomena, leading to new
insights.
\begin{itemize}
\item Section~\ref{sec:perspectives} gives an informal overview of the
  channel-based view on probability theory in terms of a number of
  perspectives. These perspectives are stated explicitly, in
  imperative form, imposing how things are --- or should be --- seen
  from a channel-based perspective, in contrast to traditional
  approaches. These perspectives will already informally use the
  notions of `state', `predicate' and `channel'.

\item Section~\ref{sec:states} commences the formal presentation of
  the ingredients of the channel-based framework, illustrating the
  concepts of state and predicate, as special forms of channels. 

\item Section~\ref{sec:conditioning} is devoted to conditioning, a key
  concept of Bayesian probability.

\item Sections~\ref{sec:channel} and~\ref{sec:bayesiannetwork} are
  devoted to channel-based Bayesian inference. First,
  Section~\ref{sec:channel} explains the use of channels as predicate
  and state transformers. Then, Section~\ref{sec:bayesiannetwork}
  illustrates this setup to model inference in a Bayesian network, for
  the standard `student' example from~\cite{KollerF09}. This section
  concludes with a clash of interpretations in an example taken
  from~\cite{Barber12}.

\item Section~\ref{sec:string} introduces a graphical calculus for
  channels --- formally justified by their definition as arrows of a
  monoidal category. The calculus encompasses and enhances the
  diagrammatic language of Bayesian networks. It offers an intuitive,
  yet mathematically rigorous, description of basic phenomena of
  probability, including conditional independency.

\item Section~\ref{sec:essence} uses the tools developed in the
  previous sections to study the relationship between joint
  distributions and their representation as Bayesian networks. First,
  we use the graphical calculus to give a channel-based account of
  disintegration. Second, we prove the equivalence of inference as
  performed on joint distributions and as performed in Bayesian
  networks (Theorem~\ref{thm:inference}). The channel perspective
  explains the different dynamics at work in the two forms of
  inference, justifying our terminology of \emph{crossover inference}
  and \emph{transformer inference} respectively.
\end{itemize}

%

\section{Perspectives}\label{sec:perspectives}

This section gives a first, informal view of the channel-based
approach, through a series of perspectives. Each of them contains a
prescription on how Bayesian phenomena appear in the channel-based
perspective, and motivates how the channel language improves more
traditional descriptions. These perspectives will informally use the
notions of `state', `predicate' and `channel'. To begin, we briefly
explain what they are. A more systematic description is given later
on.
\begin{itemize} 
\item What is usually called a discrete probability distribution, we
  call a \emph{state}. This terminology emphasises the role that
  distributions play in our programming-oriented framework: they
  express knowledge of a certain configuration (a state of affairs)
  that may be transformed by program execution (channels).


A state/distribution is represented by a convex combination of
elements from a set.  For instance, on a set $A = \{a,b,c\}$ one can
have a state $\frac{1}{3}\ket{a} + \frac{1}{2}\ket{b} +
\frac{1}{6}\ket{c}$. The `ket' notation $\ket{\cdot}$ is syntactic
sugar: it has no mathematical meaning, but echoes how states are
represented in quantum theory, where our theory may be also
instantiated~\cite{JacobsZ16}.

\item A `predicate' on a set $A$ is a function $p\colon A \rightarrow
  [0,1]$. It assigns a probability $p(a)\in [0,1]$ to each element
  $a\in A$.  Such predicates are often called `fuzzy'.  When $p(a) \in
  \{0,1\}$, so that either $p(a) = 0$ or $p(a) = 1$, for each $a\in A$
  the predicate is called sharp. A sharp predicate is traditionally called an
  event, and corresponds to a subset of $A$.

Similarly to the case of states, our terminology draws an analogy with
programming language semantics. There is a duality between states and
predicates, which goes beyond the scope of this introduction --- so
the interested reader is referred to~\cite{Jacobs17b}.

\item A `channel' $\kto{A}{B}$ from a set $A$ to another set $B$ is an
  $A$-indexed collection $\big(\omega_{a}\big)_{a\in A}$ of states
  $\omega_{a}$ on the set $B$. Alternatively, it is a function $a
  \mapsto \omega_{a}$ that sends each element $a\in A$ to a
  distribution on $B$. For $A$ and $B$ finite, yet another equivalent
  description is as a stochastic matrix with $|A|$ columns and $|B|$
  rows.
  
Channels are the pivot of our theory: states, predicates, and --- as
we shall see in Section~\ref{sec:bayesiannetwork} --- also Bayesian
networks can be seen as particular cases of a channel. More
specifically, a state $\omega$ on $B$ can be seen as a channel $f
\colon \kto{\{ \star \}}{B}$ with source the one-element set $\{ \star
\}$, defined by $f(\star) = \omega$. A predicate $p\colon A
\rightarrow [0,1]$ can be seen as a channel $\kto{A}{\{0,1\}}$ that
assigns to $a \in A$ the state $p(a)\ket{1} + (1 - p(a))\ket{0}$.
\end{itemize}

\subsection{Spell out the state}\label{subsec:states}

Our first perspective elaborates on the observation that, in
traditional probability, it is custom to leave the probability
distribution implicit, for instance in describing the probability
$\Prob(E)$ of an event $E = \{a,c\} \subseteq A$. This is justified
because this distribution, say $\omega = \frac{1}{3}\ket{a} +
\frac{1}{2}\ket{b} + \frac{1}{6}\ket{c}$, is typically fixed, so that
carrying it around explicitly, as in $\Prob_{\omega}(E)$, burdens the
notation. In contrast, in probabilistic programming, programs act on
distributions (states) and change them with every step. Hence in our
framework it makes sense to use a richer notation, where
states/distributions have a more prominent role. 

First, pursuing a more abstract, logical viewpoint, we introduce
notation $\models$ in place of $\Prob$. For an arbitrary state
$\omega$ on a set $A$ and a predicate $p\colon A \rightarrow [0,1]$ on
the same set $A$, the \emph{validity} $\omega\models p$ of $p$ in
$\omega$ is the number in $[0,1]$ given by:
\begin{equation}\label{eq:validityintro}
 \begin{array}{rcl}
\omega\models p
& \coloneqq &
\displaystyle\sum_{a\in A} \omega(a)\cdot p(a).
\end{array} 
\end{equation}

\noindent When we identify an event (sharp predicate) $E\subseteq A$
with its characteristic function $\indic{E} \colon A \rightarrow
[0,1]$, we have $\omega\models \indic{E} = \Prob_{\omega}(E) =
\frac{1}{2}$. The enhanced notation allows to distinguish this from
the probability of $E$ wrt.\ an alternative state $\psi =
\frac{1}{4}\ket{b} + \frac{3}{4}\ket{c}$, written $\psi\models
\indic{E} = \frac{3}{4}$.

Once we start treating states as explicit entities, we can give proper
attention to basic operations on states, like parallel composition
$\otimes$, marginalisation, and convex combination. These operations
will be elaborated below in Section~\ref{sec:states}.


\subsection{Conditional probability is state update with a 
predicate}\label{subsec:conditioning}

Traditionally, conditional probability is described as $\Prob(B\mid
A)$, capturing the probability of event $B$ given event $A$. This
notation is unfortunate, certainly in combination with the notation
$\Prob(B)$ for the probability of event $B$. It suggests that
conditioning $\mid$ is an operation on events, and that the
probability $\Prob(\cdot)$ of the resulting event $B\mid A$ is
computed. This perspective is sometimes called `measure-free
conditioning'~\cite{DuboisP90}. The fact that states are left
implicit, see the previous point~\ref{subsec:states}, further
contributes to the confusion.

In the view advocated here, conditioning is an operation that updates
a state $\omega$ in the light of evidence in the form of a predicate
$p$.  This is well-defined when $\omega$ and $p$ have the same
underlying set $A$, and when the validity $\omega\models p$ is
non-zero. We shall then write $\omega|_{p}$ for the state ``$\omega$
given $p$'', see Section~\ref{sec:conditioning} for more details. We
emphasise that the validity $\Prob(B\mid A)$ in state $\omega$ can now
be expressed as $\omega|_{\indic{A}} \models \indic{B}$. It is the
validity of $B$ in the state where the evidence $A$ is incorporated.

\subsection{State/predicate 
   transformation becomes explicit}\label{subsec:transformation}

The following notation $\Pr(X=a)$ often occurs in traditional
probability theory. What does it mean, and what is assumed? On close
reading we find that the following data are involved.
\begin{itemize}
\item A set, often called sample space, $\Omega$ with a
  state/distribution $\omega$ on it; please note that $\omega$ is not
  an element of $\Omega$ but a probability distribution over elements
  of $\Omega$;

\item A stochast, or random variable, $X \colon \Omega \rightarrow A$,
  for some set $A$ of outcomes;

\item An element $a\in A$ with associated event $X^{-1}(a) =
  \setin{z}{\Omega}{X(z) = a} \subseteq \Omega$;

\item The probability $\Prob(X=a)$ is then the validity of the latter
  event in the state $\omega$, that is, it is $\omega \models
  \indic{X^{-1}(a)}$.
\end{itemize}

\noindent A stochast is a special kind of channel (namely a
deterministic one). The operation $X^{-1}(a)$ will be described more
systematically as `predicate transformation' $X \ll \indic{\{a\}}$
along the channel $X$. It turns the (singleton, sharp) predicate
$\indic{\{a\}}$ on $A$ into a predicate on $\Omega$. In fact, $X \ll
\indic{\{a\}}$ can be seen as just function composition $\Omega
\rightarrow A \rightarrow [0,1]$. Since $X \ll \indic{\{a\}}$ is now a
predicate on $\Omega$, the probability $\Prob(X=a)$ can be described
more explicitly as validity: $\omega\models X \ll \indic{\{a\}}$.
More generally, for an event $E$ on $A$ we would then determine the
probability $\Prob(X\in E)$ as $\omega \models X \ll E$.

One can use the channel $X$ also for `state transformation'. In this
way one transforms the state $\omega$ on $\Omega$ into a state $X \gg
\omega$ on $A$. This operation $\gg$ is sometimes (aptly) called
pushforward, and $X\gg\omega$ is the pushforward distribution.  The
probability $\Prob(X=a)$ can equivalently be described as validity $X
\gg \omega \models \indic{\{a\}}$.

In Section~\ref{sec:channel} we elaborate on channels. One of our
findings will be that the probabilities $\omega \models c \ll p$ and
$c \gg \omega \models p$ are always the same --- for a channel $c$
from $A$ to $B$, a state $\omega$ on $A$, and a predicate
$p$ on $B$.

Moreover, we can profitably combine predicate transformation $\ll$ and
state transformation $\gg$ with conditioning of states from
point~\ref{subsec:conditioning}. As will be elaborated later on, we
can distinguish the following two basic combinations of conditioning
and transformation, with the associated terminology.
\begin{center}
\begin{tabular}{c|c|c}
\textbf{ notation } & \textbf{action} & \textbf{terminology}
\\
\hline\hline 
$\omega\big|_{c\ll q}$ 
& 
\begin{tabular}{c}
first do predicate transformation $\ll$,
\\[-0.4em]
then update the state
\end{tabular}
&
\begin{tabular}{c}
evidential reasoning, or
\\[-0.4em]
explanation, or
\\[-0.4em]
backward inference
\end{tabular}
\\
\hline
$c \gg \omega\big|_{p}$
& 
\begin{tabular}{c}
first update the state, then
\\[-0.4em]
do state transformation $\gg$,
\end{tabular}
&
\begin{tabular}{c}
causal reasoning, or
\\[-0.4em]
prediction, or
\\[-0.4em]
forward inference
\end{tabular}
\end{tabular}
\end{center}

\subsection{Use channels as probabilistic functions}\label{subsec:channel}

We have already mentioned the notation $c\colon \kto{A}{B}$ to
describe a channel $c$ from $A$ to $B$. Recall that such a channel
produces a state $c(a)$ on $B$ for each element $a\in A$. It turns out
that there is a special way to compose channels: for $c\colon
\kto{A}{B}$ and $d\colon \kto{B}{C}$ we can form a composite channel
$d \klafter c \colon \kto{A}{C}$, understood as ``$d$ after $c$''.  We
can define it via state transformation as $(d\klafter c)(a) = d \gg
c(a)$. It is not hard to check that $\klafter$ is associative, and
that there are identity maps $\idmap \colon \kto{A}{A}$, given by
$\idmap(a) = 1\ket{a}$. They form unit elements for channel
composition $\klafter$.

Abstractly, channels form morphisms in a `category'. The concept of a
category generalises the idea of sets and functions between them, to
objects and morphisms between them. These morphisms in a category need
not be actual functions, but they must be composable (and have
units). Such morphisms can be used to capture different forms of
computation, like non-deterministic, or probabilistic (via
channels). Here we shall not use categorical machinery, but use the
relevant properties in more concrete form.  For instance, composition
$\klafter$ of channels interacts appropriately with state
transformation and with predicate transformation, as in:
\[ \begin{array}{rclcrcl}
(d \klafter c) \gg \omega
& = &
d \gg (c \gg \omega)
& \mbox{\qquad and \qquad} &
(d \klafter c) \ll p
& = &
c \ll (d \ll p).
\end{array} \]

\noindent In addition to sequential composition $\klafter$ we shall
also use parallel composition $\otimes$ of channels, with an
associated calculus for combinining $\klafter$ and $\otimes$.

\subsection{Predicates are generally fuzzy}\label{subsec:fuzzy} 

In the points above we have used \emph{fuzzy} predicates, with
outcomes in the unit interval $[0,1]$, instead of the more usual
\emph{sharp} predicates, with outcomes in the two-element set
$\{0,1\}$ of Booleans. Why?
\begin{itemize}
\item The main technical reason is that fuzzy predicates are closed
  under probabilistic predicate transformation $\ll$, whereas sharp
  predicates are not. Thus, if we wish to do evidential (backward)
  reasoning $\omega|_{c\ll q}$, as described in
  point~\ref{subsec:transformation}, we are forced to use fuzzy
  predicates.

\item Fuzzy predicates are also closed under another operation, namely
  \emph{scaling}: for each $p\colon A \rightarrow [0,1]$ and $s\in
       [0,1]$ we have a new, scaled predicate $s\cdot p \colon A
       \rightarrow [0,1]$, given by $(s\cdot p)(a) = s\cdot p(a)$.
       This scaling is less important than predicate transformation,
       but still it is a useful operation.

\item Fuzzy predicates naturally fit in a probabilistic setting, where
  uncertainty is a leading concept. It thus makes sense to use this
  uncertainty also for evidence. 


\item Fuzzy predicates are simply more general than sharp predicates.
  Sharp predicates $p$ can be recognised logically among all fuzzy
  predicates via the property $p \andthen p = p$, where conjunction
  $\andthen$ is pointwise multiplication.
\end{itemize}

\noindent The traditional approach in probability theory focuses on
sharp predicates, in the form of events.  This is part of the
notation, for instance in expressions like $\Prob(X\in E)$, as used
earlier in point~\ref{subsec:conditioning}. It does not make much
sense to replace this sharp $E$ with a fuzzy $p$ when writing
$\Prob(X\in E)$. That is one more reason why we write validity via
$\models$ and not via $\Prob$. Fuzzy predicates have actually surfaced
in more recent research in Bayesian probability, see \textit{e.g.}~the
concepts of `soft' evidence~\cite{ValtortaKV02} and `uncertain'
evidence~\cite{BenMradDPLA15}, see also~\cite{Barber12}.

Fuzzy predicates have a different algebraic structure than sharp
predicates. The latter form Boolean algebras. Fuzzy predicates however
form effect modules (see \textit{e.g.}~\cite{Jacobs13a}). However,
these algebraic/logical structures will not play a role in the current
setting.

We shall later sketch how a fuzzy predicate can be replaced by an
additional node in a Bayesian network, see Remark~\ref{rem:fuzzy}.

\subsection{Marginalisation and weakening are operations}\label{subsec:margweak}

Marginalisation is the operation of turning a joint distribution
$\omega$ on a product domain $X\times Y$ into a distribution on one of
the components, say on $X$. Traditionally marginalisation is indicated
by omitting one of the variables: if $\omega(x,y)$ is written for the
joint distribution on $X\times Y$, then $\omega(x)$ is its (first)
marginal, as a distribution on $X$. It is defined as $\omega(x) =
\sum_{y} \omega(x,y)$.

We prefer to write marginalisation as an explicit operation, so that
$\fM(\omega)$ is the first marginal (on $X$), and $\sM(\omega)$ is the
second marginal (on $Y$). More generally, marginalisation can be
performed on a state $\sigma$ on a domain $X_{1}\times\cdots\times
X_{n}$ in $2^{n}$ many ways.

A seemingly different but closely related operation is weakening of
predicates. If $p\in [0,1]^{X}$ is a predicate on a domain $X$, we may
want to use it on a larger domain $X\times Y$ where we ignore the
$Y$-part. In logic this called weakening; it involves moving a
predicate to a larger context. One could also indicate this via
variables, writing $p(x)$ for the predicate on $X$, and $p(x,y)$ for
its extension to $X\times Y$, where $y$ in $p(x,y)$ is a spurious
variable. Instead we write $\fW(p) \in [0,1]^{X\times Y}$ for this
weakened predicate. It maps $(x,y)$ to $p(x)$.

Marginalisation $\margsign$ and weakening $\weaksign$ are each other's
`cousins'. As we shall see, they can both be expressed via projection
maps $\pi_{1} \colon X\times Y \rightarrow X$, namely as state
transformation $\fM(\omega) = \pi_{1} \gg \omega$ and as predicate
transformation $\fW(p) = \pi_{1} \ll p$. As a result, the symbols
$\margsign$ and $\weaksign$ can be moved accross validity $\models$,
as in~\eqref{eqn:marginalisationweakening} below. In what we call crossover
inference later on, the combination of marginalisation and weakening
plays a crucial role.

\subsection{Distinguish states and predicates}\label{subsec:distinguish}

As just argued, marginalisation is an operation on states, whereas
weakening acts on predicates (evidence). In general, certain
operations only make sense on states (like convex sum) and others on
predicates. This reflects the fact that states and predicates form
very different algebraic structures: states on a given domain form a
convex set (see \textit{e.g.}~\cite{Jacobs13a}), whereas, as already
mentioned in Section \ref{subsec:fuzzy}, predicates on a given domain
form an effect module.

Despite the important conceptual differences, states and predicates
are easily confused, also in the literature (see \textit{e.g.}~Example
\ref{ex:burglar} below). The general rule of thumb is that states
involve finitely many probabilities that add up to one --- unlike for
predicates. We elaborate formally on this distinction in
Remark~\ref{rem:statepred} below. 

On a more conceptual level, one could spell out the difference by
saying that states have an ontological flavour, whereas predicates
play an epistemological role. That means, states describe factual
reality, although in probabilistic form, via convex combinations of
combined facts. In contrast, predicates capture just the likelihoods
of individual facts as perceived by an agent. Thus probabilities in
predicates do not need to add up to one, because our perception of
reality (contrary to reality itself) is possibly inconsistent or
incomplete.\footnote{Within this perspective, it is intriguing to read
  conditioning of a state by a predicate as adapting the facts
  according to the agent's beliefs. In Philosophy one would say that
  our notion of conditioning forces an ``idealistic'' view of reality;
  in more mundane terms, it yields the possibility of ``alternative
  facts''.} We shall elaborate more on this perspective at the end of
Example~\ref{ex:burglar} below.

\section{States and predicates}\label{sec:states}

Subsection~\ref{subsec:states} claimed that states (finite probability
distributions) and fuzzy predicates --- and their different roles ---
should be given more prominence in probability theory.  We now
elaborate this point in greater detail. We thus retell the same story
as in the beginning, but this time with more mathematical details, and
with more examples.

\subsection*{States}

A \emph{state} (probability distribution) over a `sample' set $A$ is a
formal weighted combination $r_{1}\ket{a_{1}} + \cdots +
r_{n}\ket{a_{n}}$, where the $a_i$ are elements of $A$ and the $r_{i}$
are elements of $[0,1]$ with $\sum_{i}r_{i} = 1$. We shall write
$\Dst(A)$ for the set of states/distributions on a set $A$. We will
sometimes treat $\omega\in\Dst(A)$ equivalently as a `probability
mass' function $\omega \colon A \rightarrow [0,1]$ with finite support
$\supp(\omega) = \setin{a}{A}{\omega(a) \neq 0}$ and with $\sum_{a\in
  A}\omega(a) = 1$. More explicitly, the formal convex combination
$\sum_{i}r_{i}\ket{a_i}$ corresponds to the function $\omega \colon A
\rightarrow [0,1]$ with $\omega(a_{i}) = r_{i}$ and $\omega(a) = 0$ if
$a\not\in\{a_{1}, \ldots, a_{n}\}$. Then $\supp(\omega) = \{a_{1},
\ldots, a_{n}\}$, by construction.

For two states $\sigma_{1}\in\Dst(A_{1})$ and
$\sigma_{2}\in\Dst(A_{2})$, we can form the joint `product' state
$\sigma_{1}\otimes\sigma_{2} \in \Dst(A_{1}\times A_{2})$ on the cartesian
product $A_{1}\times A_{2}$ of the underlying sets, namely as:
\begin{equation}
\label{eqn:stateproduct}
\begin{array}{rcl}
(\sigma_{1}\otimes\sigma_{2})(a_{1},a_{2})
& \coloneqq &
\sigma_{1}(a_{1})\cdot\sigma_{2}(a_{2}).
\end{array}
\end{equation}

\noindent For instance, if $\sigma_{1} = \frac{1}{3}\ket{a} +
\frac{2}{3}\ket{b}$ and $\sigma_{2} = \frac{1}{8}\ket{1} +
\frac{5}{8}\ket{2} + \frac{1}{4}\ket{3}$, then their product is written with ket-notation as:
\[ \begin{array}{rcl}
\sigma_{1}\otimes\sigma_{2}
& = &
\frac{1}{24}\ket{a,1} + \frac{5}{24}\ket{a,2} + \frac{1}{12}\ket{a,3} + 
   \frac{1}{12}\ket{b,1} + \frac{5}{12}\ket{b,2} + \frac{1}{6}\ket{b,3}.
\end{array} \]

\noindent Marginalisation works in the opposite direction: it moves a
`joint' state on a product set to one of the components: for a state
$\omega\in\Dst(A_{1}\times A_{2})$ we have first and second
marginalisation $\margsign_{i}(\omega)\in\Dst(A_{i})$ determined as:
\begin{equation}
\label{eqn:marginalisation}
\begin{array}{rclcrcl}
\fM(\omega)(a_{1})
& = &
\displaystyle\sum_{a_{2}\in A_2}\omega(a_{1}, a_{2})
& \mbox{\qquad} &
\sM(\omega)(a_{2})
& = &
\displaystyle\sum_{a_{a}\in A_1}\omega(a_{1}, a_{2}).
\end{array}
\end{equation}

\noindent Here we use explicit operations $\fM$ and $\sM$ for taking
the first and second marginal. The traditional way to write a marginal
is to drop a variable: a joint distribution is written as
$\Prob(x,y)$, and its marginals as $\Prob(x)$ and $\Prob(y)$, where
$\Prob(x) = \sum_{y}P(x,y)$ and $\Prob(y) = \sum_{x}P(x,y)$.

The two original states $\sigma_{1}$ and $\sigma_{2}$ in a product
state $\sigma_{1}\otimes\sigma_{2}$ can be recovered as marginals of
this product state: $\fM(\sigma_{1}\otimes\sigma_{2}) = \sigma_{1}$
and $\sM(\sigma_{1}\otimes\sigma_{2}) = \sigma_{2}$.

In general a joint state $\omega\in\Dst(A_{1}\times A_{2})$ does
\emph{not} equal the product $\fM(\omega)\otimes\sM(\omega)$ of its
marginals, making the whole more than the sum of its parts. When we do
have $\omega = \fM(\omega)\otimes\sM(\omega)$, we call $\omega$
\emph{non-entwined}. Otherwise it is called \emph{entwined}.

\begin{example}
\label{ex:joint}
Given sets $X = \{x,y\}$ and $A = \{a,b\}$, one can prove that a state
$\omega = r_{1}\ket{x,a} + r_{2}\ket{x,b} + r_{3}\ket{y,a} +
r_{4}\ket{y,b} \in \Dst(X\times A)$, where $r_{1} + r_{2} + r_{3} +
r_{4} = 1$, is non-entwined if and only if $r_{1}\cdot r_{4} =
r_{2}\cdot r_{3}$. This fact also holds in the quantum case, see
\textit{e.g.}~\cite[\S1.5]{Mermin07}.

For instance, the following joint state is entwined:
\[ \begin{array}{rcl}
\omega
& = &
\frac{1}{8}\ket{x,a} + \frac{1}{4}\ket{x,b} + \frac{3}{8}\ket{y,a} + 
    \frac{1}{4}\ket{y,b}.
\end{array} \]

\noindent Indeed, $\omega$ has marginals $\fM(\omega)\in\Dst(X)$
and $\sM(\omega)\in\Dst(A)$, namely:
\[ \begin{array}{rclcrcl}
\fM(\omega)
& = &
\frac{3}{8}\ket{x} + \frac{5}{8}\ket{y}
& \mbox{\qquad and \qquad} &
\sM(\omega)
& = &
\frac{1}{2}\ket{a} + \frac{1}{2}\ket{b}.
\end{array} \]

\noindent The original state $\omega$ differs from the product of its
marginals:
\[ \begin{array}{rcl}
\fM(\omega)\otimes\sM(\omega)
& = &
\frac{3}{16}\ket{x,a} + \frac{3}{16}\ket{x,b} + \frac{5}{16}\ket{y,a} + 
    \frac{5}{16}\ket{y,b}.
\end{array} \]
\end{example}

There is one more operation on states that occurs frequently, namely
convex sum: if we have $n$ states $\omega_{i}\in\Dst(A)$ on the same
sets and $n$ probabilities $r_{i}\in[0,1]$ with $\sum r_{i} = 1$, then
$\sum r_{i}\omega_{i}$ is a state again.


\subsection*{Predicates}

A \emph{predicate} on a sample space (set) $A$ is a function $p \colon
A \rightarrow [0,1]$, taking values in the unit interval $[0,1]$. We
shall use the exponent notation $[0,1]^{A}$ for the set of predicates
on $A$. What in probability theory are usually called events (subsets
of $A$) can be identified with \emph{sharp} predicates, taking values
in the subset of booleans $\{0,1\} \subseteq [0,1]$. We write
$\indic{E} \in [0,1]^{A}$ for the sharp predicate associated with the
event $E \subseteq A$, defined by $\indic{E}(a) = 1$ if $a\in E$ and
$\indic{E}(a) = 0$ if $a\not\in E$. Thus predicates are a more
general, `fuzzy' notion of event, which we prefer to work with for the
reasons explained in Subsection~\ref{subsec:fuzzy}. We write $\one =
\indic{A}$, $\zero = \indic{\emptyset}$ for the truth and falsity
predicates.  They are the top and bottom elements in the set of
predicates $[0,1]^{A}$, with pointwise order. As special case, for an
element $a\in A$ we write $\indic{\{a\}}$ for the `singleton' or
`point' predicate on $A$ that is $1$ only on input $a\in A$.


For predicates $p,q\in [0,1]^{A}$ and scalar $r\in [0,1]$ we define
$p\andthen q \in [0,1]^{A}$ as $a \mapsto p(a)\cdot q(a)$ and $r\cdot
p \in [0,1]^{A}$ as $a \mapsto r\cdot p(a)$. Moreover, there is an
orthosupplement predicate $p^{\bot} \in [0,1]^{A}$ given by
$(p^{\bot})(a) = 1 - p(a)$. Then $p^{\bot\bot} = p$. Notice that
$\indic{E} \andthen \indic{D} = \indic{E \cap D}$ and
$(\indic{E})^{\bot} = \indic{\neg E}$, where $\neg E \subseteq A$ is
the set-theoretic complement of $E$.

\begin{definition}
\label{def:validity}
Let $\omega\in\Dst(A)$ be a state and $p\in [0,1]^{A}$ be a predicate,
both on the same set $A$. We write $\omega\models p$ for the
\emph{validity} or \emph{expected value} of $p$ in state
$\omega$. This validity is a number in the unit interval $[0,1]$. We
recall its definition from~\eqref{eq:validityintro}:
\begin{equation}\label{eqn:models}
\begin{array}{rcccl}
\omega\models p
& \coloneqq &
\displaystyle\sum_{a\in A} \omega(a)\cdot p(a).
\end{array}
\end{equation}
\end{definition}

For an event (sharp predicate) $E$, the probability $\Prob(E)$ wrt.\ a
state $\omega$ is defined as $\sum_{a\in E}\omega(a)$. Using
the above validity notation~\eqref{eqn:models} we write
$\omega\models\indic{E}$ instead. As special case we have
$\omega\models\indic{\{x\}} = \omega(x)$.

Notice that the validity $\omega\models\one$ of the truth predicate
$\one$ is $1$ in any state $\omega$. Similarly, $\omega\models\zero =
0$. Additionally, $\omega\models p^{\bot} = 1 - (\omega\models p)$ and
$\omega\models r\cdot p = r\cdot (\omega\models p)$.

There is also a parallel product $\otimes$ of predicates, like for
states.  Given two predicates $p_{1}\in [0,1]^{A_1}$ and $p_{2}\in
[0,1]^{A_2}$ on sets $A_{1}, A_{2}$ we form the product predicate
$p_{1}\otimes p_{2}$ on $A_{1}\times A_{2}$ via: $(p_{1}\otimes
p_{2})(a_{1},a_{2}) = p_{1}(a)\cdot p_{2}(a)$. It is not hard to see
that:
\[ \begin{array}{rcl}
\omega_{1}\otimes\omega_{2} \models p_{1}\otimes p_{2}
& = &
\big(\omega_{1}\models p_{1}\big) \cdot \big(\omega_{2}\models p_{2}\big).
\end{array} \]

\noindent A product $p\otimes\one$ or $\one\otimes p$ with the truth
predicate $\one$ corresponds to \emph{weakening}, that is to moving a
predicate $p$ to a bigger set (or context). We also write:
\begin{equation}
\label{eqn:weakening}
\begin{array}{rclcrcl}
\fW(p)
& \coloneqq &
p\otimes \one
& \mbox{\qquad and \qquad} &
\sW(p)
& \coloneqq &
\one\otimes p
\end{array}
\end{equation}

\noindent for these first and second weakening operations, like in
Subsection~\ref{subsec:margweak}. We deliberately use `dual' notation
for marginalisation $\margsign$ and weakening $\weaksign$ because
these operations are closely related, as expressed by the following
equations.
\begin{equation}
\label{eqn:marginalisationweakening}
\begin{array}{rclcrcl}
\fM(\omega) \models p
& \;=\; &
\omega \models \fW(p)
& \mbox{\qquad and \qquad} &
\sM(\omega) \models q
& \;=\; &
\omega \models \sW(q).
\end{array}
\end{equation}

\auxproof{
\[ \begin{array}{rcl}
\fM(\omega) \models p
& = &
\sum_{a} \fM(\omega)(a) \cdot p(a)
\\
& = &
\sum_{a} \big(\sum_{b}\omega(a,b)\big) \cdot p(a)
\\
& = &
\sum_{a,b} \omega(a,b) \cdot p(a)
\\
& = &
\sum_{a,b} \omega(a,b) \cdot (p\otimes\one)(a,b)
\\
& = &
\omega\models\fW(p).
\end{array} \]
}

\noindent As a result, $\sigma_{1}\otimes\sigma_{2}\models\fW(p) \,=\,
\sigma_{1}\models p$ and similarly
$\sigma_{1}\otimes\sigma_{2}\models\sW(q) \,=\, \sigma_{2}\models q$.

\begin{rem}
\label{rem:statepred}
As already mentioned in Subsection~\ref{subsec:distinguish},
conceptually, it is important to keep states and predicates apart.
They play different roles, but mathematically it is easy to confuse
them. States describe a state of affairs, whereas predicates capture
evidence. We explicitly emphasise the differences between a state
$\omega\in\Dst(A)$ and a predicate $p\colon A \rightarrow [0,1]$ in
several points.
\begin{enumerate}
\item A state has finite support. Considered as function $\omega\colon
  A\rightarrow [0,1]$, there are only finitely many elements $a\in A$
  with $\omega(a) \neq 0$. In contrast, there may be infinitely many
  elements $a\in A$ with $p(a) \neq 0$. This difference only makes
  sense when the underlying set $A$ has infinitely many elements.

\item The finite sum $\sum_{a\in A} \omega(a)$ equals $1$, since
  states involve a convex sum. In contrast there are no requirements
  about the sum of the probabilities $p(a)\in[0,1]$ for a predicate
  $p$. In fact, such a sum may not exist when $A$ is an infinite set.
  We thus see that each state $\omega$ on $A$ forms a predicate, when
  considered as a function $A \rightarrow [0,1]$. But a predicate in
  general does not form a state.

\item States and predicates are closed under completely different
  operations. As we have seen, for states we have parallel products
  $\otimes$, marginalisation $\margsign_{i}$, and convex sum. In
  contrast, predicates are closed under orthosupplement $(-)^{\bot}$,
  conjunction $\andthen$, scalar multiplication $s\cdot(-)$ and
  parallel product $\otimes$ (with weakening as special case). The
  algebraic structures of states and of predicates is completely
  different: each set of states $\Dst(A)$ forms a convex set whereas
  each set of predicates $[0,1]^{A}$ is an effect module, see
  \textit{e.g.}~\cite{Jacobs17a} for more details.

\item State transformation (along a channel) happens in a forward
  direction, whereas predicate transformation (along a channel) works
  in a backward direction. These directions are described with respect
  the direction of the channel. This will be elaborated in
  Section~\ref{sec:channel}.
\end{enumerate}
\end{rem}

\begin{rem}
\label{rem:fuzzy}
One possible reason why fuzzy predicates are not so common in
(Bayesian) probability theory is that they can be mimicked via an
extra node in a Bayesian network, together with a sharp predicate.  We
sketch how this works. Assume we have a set $X = \{a,b,c\}$ and we
wish to consider a fuzzy predicate $p\colon X \rightarrow [0,1]$ on
$X$, say with $p(a) = \frac{2}{3}$, $p(b) = \frac{1}{2}$ and $p(c) =
\frac{1}{4}$. Then we can introduce an extra node $2 = \{t,f\}$ with a
channel $h\colon \kto{A}{2}$ given by:
\[ \begin{array}{rclcrclcrcl}
h(a)
& = &
\frac{2}{3}\ket{t} + \frac{1}{3}\ket{f}
& \mbox{\qquad} &
h(b)
& = &
\frac{1}{2}\ket{t} + \frac{1}{2}\ket{f}
& \mbox{\qquad} &
h(c)
& = &
\frac{1}{4}\ket{t} + \frac{3}{4}\ket{f}.
\end{array} \]

\noindent The original predicate $p$ on $X = \{a,b,c\}$ can now be
reconstructed via predicate transformation along $h$ as $h \ll
\indic{\{t\}}$, where, recall, $\indic{\{t\}}$ is the sharp predicate
on $2$ which is $1$ at $t$ and $0$ at $f$.

As an aside: we have spelled out the general isomorphism between
predicates on a set $A$ and channels $\kto{A}{2}$. Conceptually this
is pleasant, but in practice we do not wish to extend our Bayesian
network every time a fuzzy predicate pops up. What this example also
illustrates is that sharpness of predicates is not closed under
predicate transformation.
\end{rem}

\section{Conditioning}\label{sec:conditioning}

Conditioning is one of the most fundamental operations in probability
theory. It is the operation that updates a state in the light of
certain evidence. This evidence is thus incorporated in a new, updated
state, that reflects the new insight. For this reason conditioning is
sometime called belief update or belief revision. It forms the basis
of learning, training and inference, see also
Section~\ref{sec:bayesiannetwork}.

A conditional probability is usually written as $\Prob(E\mid D)$. It
describes the probability of event $E$, given event $D$. In the
current context we follow a more general path, using fuzzy predicates
instead of events. Also, we explicitly carry the state around. From
this perspective, the update of a state $\omega$ with a predicate $p$,
leading to an updated state $\omega|_{p}$, is the fundamental
operation. It allows us to retrieve probabilities $\Pr(E\mid D)$ as
special case, as will be shown at the end of this section.

\begin{definition}
\label{def:conditioning}
Let $\omega\in\Dst(A)$ be a state and $p\in [0,1]^{A}$ be a predicate,
both on the same set $A$. If the validity $\omega\models p$ is
non-zero, we write $\omega|_{p}$ or for the conditional state
``$\omega$ given $p$'', defined as formal convex sum:
\begin{equation}
\label{eqn:conditioning}
\begin{array}{rcl}
\omega|_{p}
& \df &
\displaystyle \sum_{a\in A}
   \frac{\omega(a)\cdot p(a)}{\omega\models p}\bigket{a}.
\end{array}
\end{equation}
\end{definition}

\begin{example}
\label{ex:conditioning}
Let's take the numbers of a dice as sample space: $\textsf{pips} =
\{1,2,3,4,5,6\}$, with a fair/uniform dice distribution $\textsf{dice}
\in\Dst(\textsf{pips})$.
\[ \begin{array}{rcl}
\textsf{dice}
& = &
\frac{1}{6}\ket{1} + \frac{1}{6}\ket{2} + \frac{1}{6}\ket{3} + 
   \frac{1}{6}\ket{4} + \frac{1}{6}\ket{5} + \frac{1}{6}\ket{6}.
\end{array} \]

\noindent We consider the predicate $\textsf{evenish} \in [0,1]^{\textsf{pips}}$
expressing that we are fairly certain of pips being even:
\[ \begin{array}{rclcrclcrcl}
\textsf{evenish}(1)
& = &
\frac{1}{5}
& \mbox{\qquad} &
\textsf{evenish}(3)
& = &
\frac{1}{10}
& \mbox{\qquad} &
\textsf{evenish}(5)
& = &
\frac{1}{10}
\\
\textsf{evenish}(2)
& = &
\frac{9}{10}
& &
\textsf{evenish}(4)
& = &
\frac{9}{10}
& &
\textsf{evenish}(6)
& = &
\frac{4}{5}
\end{array} \]

\noindent We first compute the validity of $\textsf{evenish}$ for our fair
dice:
\[ \begin{array}{rcl}
\textsf{dice}\models\textsf{evenish}
& = &
\sum_{x} \textsf{dice}(x)\cdot\textsf{evenish}(x)
\\
& = &
\frac{1}{6}\cdot\frac{1}{5} + \frac{1}{6}\cdot\frac{9}{10} + 
   \frac{1}{6}\cdot\frac{1}{10} + \frac{1}{6}\cdot\frac{9}{10} + 
   \frac{1}{6}\cdot\frac{1}{10} + \frac{1}{6}\cdot\frac{4}{5} 
\hspace*{\arraycolsep}=\hspace*{\arraycolsep}
\frac{2 + 9 + 1 + 9 + 1 + 8}{60}
\hspace*{\arraycolsep}=\hspace*{\arraycolsep}
\frac{1}{2}.
\end{array} \]

\noindent If we take \textsf{evenish} as evidence, we can update our
state and get:
\[ \begin{array}{rcl}
\textsf{dice}\big|_{\textsf{evenish}}
& = &
\displaystyle\sum_{x}\frac{\textsf{dice}(x)\cdot\textsf{evenish}(x)}
   {\textsf{dice}\models\textsf{evenish}}\bigket{x}
\\
& = &
\frac{\nicefrac{1}{6}\cdot\nicefrac{1}{5}}{\nicefrac{1}{2}}\ket{1} + 
   \frac{\nicefrac{1}{6}\cdot\nicefrac{9}{10}}{\nicefrac{1}{2}}\ket{2} + 
   \frac{\nicefrac{1}{6}\cdot\nicefrac{1}{10}}{\nicefrac{1}{2}}\ket{3} + 
   \frac{\nicefrac{1}{6}\cdot\nicefrac{9}{10}}{\nicefrac{1}{2}}\ket{4} + 
   \frac{\nicefrac{1}{6}\cdot\nicefrac{1}{10}}{\nicefrac{1}{2}}\ket{5} + 
   \frac{\nicefrac{1}{6}\cdot\nicefrac{4}{5}}{\nicefrac{1}{2}}\ket{6}
\\
& = &
\frac{1}{15}\ket{1} + \frac{3}{10}\ket{2} + \frac{1}{30}\ket{3} + 
   \frac{3}{10}\ket{4} + \frac{1}{30}\ket{5} + \frac{4}{15}\ket{6}.
\end{array} \]

\noindent As expected, the probability of the even pips is now higher
than the odd ones.
\end{example}

We collect some basic properties of conditioning.

\begin{lemma}
\label{lem:conditioning}
Let $\omega\in\Dst(A)$ and $p,q\in [0,1]^{A}$ be a state with
predicates on the same set $A$.
\begin{enumerate}
\item \label{lem:conditioning:one} Conditioning with
truth does nothing: $\omega|_{\one} = \omega$;

\item \label{lem:conditioning:andthen} Conditioning with a conjunction
  amounts to separate conditionings: $\omega|_{p\andthen q} =
  \big(\omega|_{p}\big)\big|_{q}$;

\item \label{lem:conditioning:scalar} Conditioning with scalar product
  has no effect, when the scalar is non-zero: $\omega|_{r\cdot p} =
  \omega|_{p}$ when $r\neq 0$;

\item \label{lem:conditioning:point} Conditioning with a point
  predicate yields a point state: $\omega|_{\indic{\{x\}}} = 1\ket{x}$,
  when $\omega(x) \neq 0$;
\end{enumerate}

\noindent Now let $\sigma_{i}\in\Dst(A_{i})$ and $p_{i}\in[0,1]^{A_i}$.

\begin{enumerate}[resume]
\item \label{lem:conditioning:product}
  $(\sigma_{1}\otimes\sigma_{2})\big|_{p_{1}\otimes p_{2}} =
  (\sigma_{1}|_{p_1})\otimes(\sigma_{2}|_{p_2})$.

\item \label{lem:conditioning:marginal}
$\fM\big((\sigma\otimes\tau)|_{\subfW(p_{1})}\big) = \sigma|_{p_1}$ and
$\sM\big((\sigma\otimes\tau)|_{\subsW(p_{2})}\big) = \tau|_{p_{2}}$.
\end{enumerate}
\end{lemma}

\begin{myproof}
All these properties follow via straightforward computation. We shall
do~\eqref{lem:conditioning:andthen}
and~\eqref{lem:conditioning:product}.

For~\eqref{lem:conditioning:andthen} we use:
\[ \begin{array}{rcl}
\big((\omega|_{p})|_{q}\big)(a)
\hspace*{\arraycolsep}=\hspace*{\arraycolsep}
\displaystyle \frac{\omega|_{p}(a)\cdot q(a)}{\omega|_{p}\models q}
\hspace*{\arraycolsep}=\hspace*{\arraycolsep}
\displaystyle \frac{\frac{\omega(a)\cdot p(a)}{\omega\models p}\cdot q(a)}
    {\sum_{b}\omega|_{p}(b)\cdot q(b)}
& = &
\displaystyle \frac{\frac{\omega(a)\cdot p(a)}{\omega\models p}\cdot q(a)}
    {\sum_{b}\frac{\omega(b)\cdot p(b)}{\omega\models p}\cdot q(b)}
\\[+1.3em]
& = &
\displaystyle \frac{\omega(a) \cdot p(a) \cdot q(a)}
   {\sum_{b}\omega(b) \cdot p(b)\cdot q(b)}
\\[+0.8em]
& = &
\displaystyle \frac{\omega(a) \cdot (p\andthen q)(a)}
   {\omega\models p\andthen q}
\\
& = &
\big(\omega|_{p\andthen q}\big)(a).
\end{array} \]

\noindent Similarly, for~\eqref{lem:conditioning:product} we use:
$$\begin{array}[b]{rcl}
(\sigma_{1}\otimes\sigma_{2})\big|_{p_{1}\otimes p_{2}}(a_{1},a_{2})
& = &
\displaystyle\frac{(\sigma_{1}\otimes\sigma_{2})(a_{1},a_{2})
   \cdot (p_{1}\otimes p_{2})(a_{1},a_{2})}
   {(\sigma_{1}\otimes\sigma_{2})\models (p_{1}\otimes p_{2})}
\\[+1em]
& = &
\displaystyle\frac{\sigma_{1}(a_{1}) \cdot \sigma_{2}(a_{2}) \cdot
    p_{1}(a_{1}) \cdot p_{2}(a)}
    {(\sigma_{1}\models p_{1})\cdot(\sigma_{2}\models p_{2})}
\\[+1em]
& = &
\displaystyle\frac{\sigma_{1}(a_{1}) \cdot p_{1}(a_{1})}{\sigma_{1}\models p_{1}}
\cdot \frac{\sigma_{2}(a_{2}) \cdot p_{2}(a)}{\sigma_{2}\models p_{2}}
\\[+0.5em]
& = &
(\sigma_{1}|_{p_1})\otimes(\sigma_{2}|_{p_2}).
\end{array} \eqno{\QEDbox}$$
\end{myproof}

The following result gives the generalisation of Bayes' rule to the
current setting with states and predicates.

\begin{theorem}
\label{thm:bayes}
Let $\omega\in\Dst(A)$ and $p,q\in [0,1]^{A}$ be a state and two predicates
on the set $A$.
\begin{enumerate}
\item \label{thm:bayes:product} The \emph{product rule} holds:
\begin{equation}
\label{eqn:product}
\begin{array}{rcl}
\omega|_{p} \models q
& \;=\; &
\displaystyle\frac{\omega\models p\andthen q}{\omega\models p}
\end{array}
\end{equation}

\item \label{thm:bayes:bayes} \emph{Bayes' rule} holds:
\begin{equation}
\label{eqn:bayes}
\begin{array}{rcl}
\omega|_{p} \models q
& \;=\; &
\displaystyle\frac{(\omega|_{q}\models p)\cdot(\omega\models q)}
   {\omega\models p}
\end{array}
\end{equation}
\end{enumerate}
\end{theorem}

\begin{myproof}
Point~\eqref{thm:bayes:bayes} follows directly
from~\eqref{thm:bayes:product} by using that $p\andthen q = q\andthen
p$, so we concentrate on~\eqref{thm:bayes:product}. 
$$\begin{array}[b]{rcl}
\omega|_{p} \models q
\hspace*{\arraycolsep}=\hspace*{\arraycolsep}
\displaystyle\sum_{a} \omega|_{p}(a)\cdot q(a)
\hspace*{\arraycolsep}=\hspace*{\arraycolsep}
\displaystyle\sum_{a}\frac{\omega(a)\cdot p(a)}{\omega\models p}\cdot q(a)
& = &
\displaystyle\frac{\sum_{a}\omega(a)\cdot (p\andthen q)(a)}{\omega\models p}
\\
& = &
\displaystyle\frac{\omega\models p\andthen q}{\omega\models p}.
\end{array} \eqno{\QEDbox}$$
\end{myproof}

We now relate our state-and-predicate based approach to conditioning
to the traditional one. Recall that for events $E,D\subseteq A$ one
has, by definition:
\[ \begin{array}{rcl}
\Pr(E\mid D)
& = &
\displaystyle\frac{\Prob(E\cap D)}{\Prob(D)}.
\end{array} \]

\noindent If these probabilities $\Prob(\cdot)$ are computed wrt.\ a
distribution $\omega\in\Dst(A)$, we can continue as follows.
\[ \begin{array}{rcccccccl}
\Pr(E\mid D)
& = &
\displaystyle\frac{\Prob(E\cap D)}{\Prob(D)}
& = &
\displaystyle\frac{\omega\models\indic{E\cap D}}{\omega\models\indic{D}}
& = &
\displaystyle\frac{\omega\models\indic{E}\andthen\indic{D}}
   {\omega\models\indic{D}}
& \smash{\stackrel{\eqref{eqn:product}}{=}} &
\omega|_{\indic{D}}\models\indic{E}.
\end{array} \]

\noindent Thus the probability $\Prob(E\mid D)$ can be expressed in
our framework as the validity of the sharp predicate $E$ in the state
updated with the sharp predicate $D$. This is precisely the intended
meaning.

\section{Bayesian inference via state/predicate 
   transformation}\label{sec:channel}

As mentioned in Subsection~\ref{subsec:channel}, a channel $c\colon
\kto{A}{B}$ between two sets $A,B$ is a probabilistic function from
$A$ to $B$. It maps an an element $a\in A$ to a state $c(a)\in\Dst(B)$
of $B$. Hence it is an actual function of the form $A \rightarrow
\Dst(B)$. Such functions are often described as conditional
probabilities $a \mapsto \Prob(b\mid a)$, or as stochastic
matrices. We repeat that channels are fundamental --- more so than
states and predicates --- since a state $\omega\in\Dst(A)$ can be
identified with a channel $\omega\colon \kto{1}{A}$ for the singleton
set $1 = \{0\}$. Similarly, a predicate $p\in [0,1]^{A}$ can be
identified with a channel $p\colon \kto{A}{2}$, where $2 = \{0,1\}$;
this uses that $\Dst(2) \cong [0,1]$.

Channels are used for probabilistic state transformation $\gg$ and
predicate transformation $\ll$, in the following manner.

\begin{definition}
\label{def:transformation}
Let $c\colon \kto{A}{B}$ be a channel, with a state $\omega\in\Dst(A)$
on its domain $A$ and a predicate $q\in [0,1]^{B}$ on its codomain
$B$.
\begin{enumerate}
\item State transformation yields a state $c \gg \omega$ on $B$ defined by:
\begin{equation}
\label{eqn:stattransf}
\begin{array}{rcl}
\big(c \gg \omega\big)(b)
& \coloneqq &
\displaystyle\sum_{a\in A} \omega(a)\cdot c(a)(b).
\end{array}
\end{equation}

\item Predicate transformation gives a predicate $c \ll q$ on $A$
  defined by:
\begin{equation}
\label{eqn:predtransf}
\begin{array}{rcl}
\big(c \ll q\big)(a)
& \coloneqq &
\displaystyle\sum_{b\in B} c(a)(b) \cdot q(b).
\end{array}
\end{equation}
\end{enumerate}
\end{definition}

The next example illustrates how state and predicate transformation
can be used systematically to reason about probabilistic questions.

\begin{example}
\label{ex:transformation}
In a medical context we distinguish patients with low (L), medium (M),
and high (H) blood pressure. We thus use as `blood' sample space $B =
\{L,M,H\}$, say with initial (`prior' or `base rate') distribution
$\beta\in\Dst(B)$:
\[ \begin{array}{rcl}
\beta
& = &
\frac{1}{8}\ket{L} + \frac{1}{2}\ket{M} + \frac{3}{8}\ket{H}.
\end{array} \]

\noindent We consider a particular disease, whose a priori occurrence
in the population depends on the blood pressure, as given by the
following table.
\begin{center}
\begin{tabular}{c|c}
\textbf{blood pressure} & \textbf{disease likelihood}
\\
\hline\hline
Low & 5\%
\\
\hline
Medium & 10\%
\\
\hline
High & 15\%
\end{tabular}
\end{center}

\noindent We choose as sample space for the disease $D = \{d,
d^{\bot}\}$ where the element $d$ represents presence of the disease
and $d^{\bot}$ represents absence. The above table is now naturally
described as a `sickness' channel $s \colon \kto{B}{D}$, given by:
\[ \begin{array}{rclcrclcrcl}
s(L)
& = &
0.05\ket{d} + 0.95\ket{d^\bot}
& \mbox{\quad} &
s(M)
& = &
0.1\ket{d} + 0.9\ket{d^\bot}
& \mbox{\quad} &
s(H)
& = &
0.15\ket{d} + 0.85\ket{d^\bot}.
\end{array} \]

\noindent We ask ourselves two basic questions.
\begin{enumerate}
\item \textbf{What is the a priori probability of the disease?}  The
  answer to this question is obtained by state transformation, namely
  by transforming the blood pressure distribution $\beta$ on $B$ to a
  disease distribution $s \gg \beta$ on $D$ along the sickness channel
  $s$. Concretely:
\[ \begin{array}{rcl}
\big(s \gg \beta)(d)
& \smash{\stackrel{\eqref{eqn:stattransf}}{=}} &
\sum_{x\in B} \beta(x)\cdot s(x)(d)
\\
& = &
\beta(L)\cdot s(L)(d) + \beta(M)\cdot s(M)(d) + \beta(H)\cdot s(H)(d) 
\\
& = &
\frac{1}{8}\cdot\frac{1}{20} + \frac{1}{2}\cdot\frac{1}{10} +
   \frac{3}{8}\cdot\frac{3}{20}
\\
& = &  
\frac{9}{80}
\\
\big(s \gg \beta)(d^{\bot})
& \smash{\stackrel{\eqref{eqn:stattransf}}{=}} &
\sum_{x\in B} \beta(x)\cdot s(x)(d^{\bot})
\\
& = &
\beta(L)\cdot s(L)(d^{\bot}) + \beta(M)\cdot s(M)(d^{\bot}) + 
    \beta(H)\cdot s(H)(d^{\bot}) 
\\
& = &
\frac{1}{8}\cdot\frac{19}{20} + \frac{1}{2}\cdot\frac{9}{10} +
   \frac{3}{8}\cdot\frac{17}{20}
\\
& = & 
\frac{71}{80}.
\end{array} \]

\noindent Thus we obtain as a priori disease distribution $c \gg \beta
= \frac{9}{80}\ket{d} + \frac{71}{80}\ket{d^{\bot}} = 0.1125\ket{d} +
0.8875\ket{d^{\bot}}$. A bit more than 11\% of the population has the
disease at hand.

\medskip

\item \textbf{What is the likely blood pressure for people without the
  disease?}  Before we calculate the updated (`a posteriori') blood
  pressure distribution, we reason intuitively. Since non-occurrence
  of the disease is most likely for people with low blood pressure, we
  expect that the updated blood pressure --- after taking the evidence
  `absence of disease' into account --- will have a higher probability
  of low blood pressure than the orignal (a priori) value of
  $\frac{1}{8}$ in $\beta$.

The evidence that we have is the point predicate $\indic{\{d^\bot\}}$
on $D$, representing absence of the disease. In order to update
$\beta\in\Dst(B)$ we first apply predicate transformation $s \ll
\indic{\{d^\bot\}}$ to obtain a predicate on $B$. This transformed
predicate in $[0,1]^{B}$ is computed as follows.
\[ \begin{array}{rcl}
\big(s \ll \indic{\{d^\bot\}})(L)
& \smash{\stackrel{\eqref{eqn:predtransf}}{=}} &
\sum_{y\in D} s(L)(y) \cdot \indic{\{d^\bot\}}(y)
\hspace*{\arraycolsep}=\hspace*{\arraycolsep}
s(L)(d^{\bot})
\hspace*{\arraycolsep}=\hspace*{\arraycolsep}
0.95
\\
\big(s \ll \indic{\{d^\bot\}})(M)
& \smash{\stackrel{\eqref{eqn:predtransf}}{=}} &
\sum_{y\in D} s(M)(y) \cdot \indic{\{d^\bot\}}(y)
\hspace*{\arraycolsep}=\hspace*{\arraycolsep}
s(M)(d^{\bot})
\hspace*{\arraycolsep}=\hspace*{\arraycolsep}
0.9
\\
\big(s \ll \indic{\{d^\bot\}})(H)
& \smash{\stackrel{\eqref{eqn:predtransf}}{=}} &
\sum_{y\in D} s(H)(y) \cdot \indic{\{d^\bot\}}(y)
\hspace*{\arraycolsep}=\hspace*{\arraycolsep}
s(H)(d^{\bot})
\hspace*{\arraycolsep}=\hspace*{\arraycolsep}
0.85.
\end{array} \]

\noindent Notice that although $\indic{\{d^\bot\}}$ is a sharp
predicate, the transformed predicate $s \ll \indic{\{d^\bot\}}$ is not
sharp. This shows that sharp predicates are not closed under predicate
transformation --- as mentioned earlier in
Subsection~\ref{subsec:fuzzy}.

We can now update the original blood pressure distribution $\beta$
with the transformed evicence $s \ll \indic{\{d^\bot\}}$. We first
compute validity, and then perform conditioning:
\[ \begin{array}{rcl}
\beta \models s \ll \indic{\{d^\bot\}}
& \smash{\stackrel{\eqref{eqn:models}}{=}} &
\sum_{x\in B} \beta(x) \cdot (s \ll \indic{\{d^\bot\}})(x)
\\
& = &
\beta(L) \cdot (s \ll \indic{\{d^\bot\}})(L) + 
   \beta(M) \cdot (s \ll \indic{\{d^\bot\}})(M) + 
   \beta(H) \cdot (s \ll \indic{\{d^\bot\}})(H)
\\
& = &
\frac{1}{8}\cdot\frac{19}{20} + \frac{1}{2}\cdot\frac{9}{10} +
   \frac{3}{8}\cdot\frac{17}{20}
\\
& = &
\frac{71}{80}
\\
\beta\big|_{s \ll \indic{\{d^\bot\}}}
& \smash{\stackrel{\eqref{eqn:conditioning}}{=}} &
\displaystyle\sum_{x\in B} \frac{\beta(x)\cdot (s \ll \indic{\{d^\bot\}})(x)}
    {\beta \models s \ll \indic{\{d^\bot\}}}\bigket{x}
\\[+1em]
& = &
\displaystyle
\frac{\nicefrac{1}{8}\cdot\nicefrac{19}{20}}{\nicefrac{71}{80}}\ket{L} + 
   \frac{\nicefrac{1}{2}\cdot\nicefrac{9}{10}}{\nicefrac{71}{80}}\ket{M} +
   \frac{\nicefrac{3}{8}\cdot\nicefrac{17}{20}}{\nicefrac{71}{80}}\ket{H}
\\
& = &
\frac{19}{142}\ket{L} + \frac{36}{71}\ket{M} + \frac{51}{142}\ket{H}
\\
& \sim &
0.134\ket{L} + 0.507\ket{M} + 0.359\ket{H}.
\end{array} \]

\noindent As intuitively expected, a posteriori the probability of low
blood pressure is higher than in the a priori distribution $\beta$ ---
and the probability of high blood pressure is lower too.
\end{enumerate}

\noindent These calculations with probabilities are relatively easy
but may grow out of hand quickly. Therefore a library has been
developed, called \EfProb see~\cite{ChoJ17b}, that provides the
relevant functions, for validity, state update, state and predicate
transformation, \textit{etc.}

It is natural to see a state $\beta$ and a channel $s$, as used above,
as stochastic matrices $M_{\beta}$ and $M_{s}$, of the form:
\[ \begin{array}{rclcrcl}
M_{\beta}
& = &
\left(\begin{matrix}
\nicefrac{3}{8}
\\
\nicefrac{1}{2}
\\
\nicefrac{3}{8}
\end{matrix}\right)
& \mbox{\qquad} &
M_{s}
& = &
\left(\begin{matrix}
0.05 & 0.1 & 0.15
\\
0.95 & 0.9 & 0.85
\end{matrix}\right)
\end{array} \]

\noindent These matrices are called stochastic because the columns add
up to 1. The matrix of the state $s \gg \beta$ is then obtained by
matrix multiplication $M_{s}M_{\beta}$. For predicate transformation
$s \ll \indic{\{d^\bot\}}$ with $M_{\indic{\{d^\bot\}}} =
\left(\begin{matrix} 0 & 1 \end{matrix}\right)$ one uses matrix 
multiplication in a different order:
\[ \begin{array}{rcccl}
M_{\indic{\{d^\bot\}}}M_{s}
& = &
\left(\begin{matrix} 0 & 1 \end{matrix}\right)
\left(\begin{matrix}
0.05 & 0.1 & 0.15
\\
0.95 & 0.9 & 0.85
\end{matrix}\right)
& = &
\left(\begin{matrix}
0.95 & 0.9 & 0.85
\end{matrix}\right).
\end{array} \]

\end{example}

The diligent reader may have noticed in this example that the
probability $(s \gg \beta)(d^{\bot}) = (s \gg \beta) \models
\indic{\{d^\bot\}} = \frac{71}{80}$ in Example~\ref{ex:transformation}
coincides with the probability $\beta \models (s \ll
\indic{\{d^\bot\}}) = \frac{71}{80}$. This in fact in an instance of
the following general result, relating validity and transformations.

\begin{proposition}
\label{prop:validitytransformation}
Let $c\colon \kto{A}{B}$ be a channel, $\omega\in\Dst(A)$ be a state
on its domain, and $q\in [0,1]^{B}$ a predicate on its codomain. Then:
\begin{equation}
\label{eqn:validitytransformation}
\begin{array}{rcl}
(c \gg \omega) \models q
& \;=\; &
\omega \models (c \ll q).
\end{array}
\end{equation}
\end{proposition}

\begin{myproof}
The result follows from a simple calculation:
$$\begin{array}[b]{rcl}
(c \gg \omega) \models q
& \smash{\stackrel{\eqref{eqn:models}}{=}} &
\displaystyle\sum_{b\in B} (c\gg \omega)(b)\cdot q(b)
\\
& \smash{\stackrel{\eqref{eqn:stattransf}}{=}} &
\displaystyle\sum_{b\in B} 
   \left(\sum_{a\in A} \omega(a)\cdot c(a)(b)\right)\cdot q(b)
\\
& = &
\displaystyle\sum_{a\in A, b\in B} \omega(a) \cdot c(a)(b) \cdot q(b)
\\
& = &
\displaystyle\sum_{a\in A} \omega(a) \cdot 
   \left(\sum_{b\in B}c(a)(b) \cdot q(b)\right)
\\
& \smash{\stackrel{\eqref{eqn:predtransf}}{=}} &
\displaystyle\sum_{a\in A} \omega(a) \cdot (c \ll q)(a)
\\
& \smash{\stackrel{\eqref{eqn:models}}{=}} &
\omega \models (c \ll q).
\end{array} \eqno{\QEDbox}$$
\end{myproof}

There are two more operations on channels that we need to consider,
namely sequential composition $\klafter$ and parallel composition
$\otimes$.

\begin{definition}
\label{def:chancomposition}
Consider channels $f\colon \kto{A}{B}$, $g\colon \kto{C}{D}$ and 
$h\colon \kto{X}{Y}$. These channels can be composed sequentially
and in parallel, yielding new channels:
\[ g \klafter f \colon \kto{A}{C}
\mbox{\qquad and \qquad}
f\otimes h \colon \kto{A\times X}{B\times Y}, \]

\noindent via the following definitions.
\[ \begin{array}{rclcrcl}
(g \klafter f)(a)
& \coloneqq &
g \gg f(a)
& \mbox{\qquad so that \qquad} &
(g \klafter f)(a)(c)
& = &
\displaystyle\sum_{b\in B} f(a)(b)\cdot g(b)(c).
\end{array} \]

\noindent The latter formula shows that channel composition is
essentially matrix multiplication.

Next,
\[ \begin{array}{rclcrcl}
(f\otimes h)(a,x)
& \coloneqq &
f(a) \otimes h(x)
& \mbox{\qquad so that \qquad} &
(f\otimes h)(a,x)(b,y)
& = &
f(a)(b) \cdot h(x)(y).
\end{array} \]

\noindent The product $\otimes$ on the right of $\coloneqq$ is the
product of states, as described in~\eqref{eqn:stateproduct}. In terms
of matrices, parallel composition of channels is given by the
Kronecker product.
\end{definition}

It is not hard to see that $\klafter$ and $\otimes$ are well-behaved
operations, satisfying for instance:
\[ \begin{array}{rcl}
\big(g\otimes k\big) \klafter \big(f\otimes h\big) 
& = &
\big(g \klafter f\big) \otimes \big(k \klafter h\big).
\end{array} \]

\noindent They interact nicely with state and predicate transformation:
\[ \begin{array}{rclcrcl}
\big(g \klafter f) \gg \omega
& = &
g \gg \big(f \gg \omega)
& \mbox{\qquad} &
\big(g \klafter f) \ll q
& = &
f \ll \big(g \ll q)
\\
\big(f\otimes h) \gg (\sigma\otimes\tau)
& = &
\big(f \gg \sigma\big) \otimes \big(h \gg \tau\big)
& & 
\big(f\otimes h) \ll (p\otimes q)
& = &
\big(f \ll p\big) \otimes \big(g \ll q\big).
\end{array} \]

\noindent Moreover, for the identity channel $\idmap$ given by $\idmap(x)
= 1\ket{x}$ we have:
\[ \begin{array}{rccclcrcl}
\idmap \klafter f
& = &
f
& = &
f \klafter \idmap
& \mbox{\qquad} &
\idmap\otimes\idmap
& = &
\idmap.
\end{array} \]

\noindent We will see examples of parallel composition of channels in
Section~\ref{sec:string} when we discuss (the semantics of) Bayesian
networks.

\begin{rem}
\label{rem:deterministichannel}
An ordinary function $f\colon A \rightarrow B$ can be turned into
a `deterministic' channel $\klin{f} \colon \kto{A}{B}$ via:
\begin{equation}
\label{eqn:deterministichannel}
\begin{array}{rcl}
\klin{f}(a)
& \coloneqq &
1\ket{f(a)}.
\end{array}
\end{equation}

\noindent This operation $\klin{\cdot}$ sends function composition to
channel composition: $\klin{g \after f} = \klin{g} \klafter \klin{f}$.
The random variable $X\colon \Omega \rightarrow A$ that we used in
Subsection~\ref{subsec:transformation} is an example of such a
deterministic channel. Formally, we should now write $X^{-1}(a) =
\klin{X} \ll \indic{\{a\}}$ for the event $X^{-1}(a)$ on $\Omega$.

There are some further special cases of deterministic channels that we
mention explicitly.
\begin{enumerate}
\item For two sets $A_{1},A_{2}$ we can form the cartesian product
  $A_{1}\times A_{2}$ with its two projection functions $\pi_{1}
  \colon A_{1}\times A_{2} \rightarrow A_{1}$ and $\pi_{2} \colon
  A_{1}\times A_{2} \rightarrow A_{2}$. They can be turned into
  (deterministic) channels $\klin{\pi_{i}} \colon \kto{A_{1}\times
    A_{2}}{A_{i}}$. One can then see that marginalisation and
  weakening are state transformation and predicate transformation
  along these projection channels:
\[ \begin{array}{rclcrcl}
\klin{\pi_{1}} \gg \omega
& = &
\fM(\omega)
& \mbox{\qquad} &
\klin{\pi_{1}} \ll p
& = &
\fW(p)
\hspace*{\arraycolsep}=\hspace*{\arraycolsep}
p\otimes\one
\\
\klin{\pi_{2}} \gg \omega
& = &
\sM(\omega)
& \mbox{\qquad} &
\klin{\pi_{2}} \ll q
& = &
\sW(q)
\hspace*{\arraycolsep}=\hspace*{\arraycolsep}
\one\otimes q
\end{array} \]

\noindent As a result, equation~\eqref{eqn:marginalisationweakening}
is a special case of~\eqref{eqn:validitytransformation}.

Moreover, these projection channels commute with parallel
composition $\otimes$ of channels, in the sense that:
\[ \begin{array}{rclcrcl}
\klin{\pi_1} \klafter \big(f \otimes h\big)
& = &
f \klafter \klin{\pi_1}
& \mbox{\qquad} &
\klin{\pi_2} \klafter \big(f \otimes h\big)
& = &
h \klafter \klin{\pi_2}
\end{array} \]

\item For each set $A$ there is a diagonal (or `copy') function
  $\Delta \colon A \rightarrow A\times A$ with $\Delta(a) = (a,a)$.
  It can be turned into a channel too, as $\klin{\Delta} \colon
  \kto{A}{A\times A}$. However, this copy channel does \emph{not}
  interact well with parallel composition of channels, in the sense
  that in general:
\[ \begin{array}{rcl}
\klin{\Delta} \klafter f
& \neq &
\big(f\otimes f\big) \klafter \klin{\Delta}.
\end{array} \]

\noindent This equation does hold when the channel $f$ is
deterministic.  Via diagonals we can relate parallel products
$\otimes$ and conjunctions $\andthen$ of predicates:
\[ \begin{array}{rcl}
\klin{\Delta} \ll \big(p\otimes q\big)
& = &
p \andthen q.
\end{array} \]

\auxproof{
\[ \begin{array}{rcl}
\big(\klin{\Delta} \ll \big(p\otimes q\big)\big)(a)
& = &
\sum_{x,y} \klin{\Delta}(a)(x,y) \cdot (p\otimes q)(x,y)
\\
& = &
(p \otimes q)(a,a)
\\
& = &
p(a)\cdot q(a)
\\
& = &
(p \andthen q)(a).
\end{array} \]
}
\end{enumerate}

\noindent In the sequel we often omit the braces $\klin{\cdot}$ around
projections and diagonals, and simply write projection and copy
channels as $\pi_{i} \colon \kto{A_{1}\times A_{2}}{A_{i}}$ and
$\Delta \colon \kto{A}{A\times A}$.
\end{rem}

\section{Inference in Bayesian networks}\label{sec:bayesiannetwork}

In this section we illustrate how channels can be used both
to model Bayesian networks and to reason about them. We
shall use a standard example from the literature, namely the `student'
network from~\cite{KollerF09}. 
\begin{figure}
\begin{center}
\includegraphics[scale=0.5]{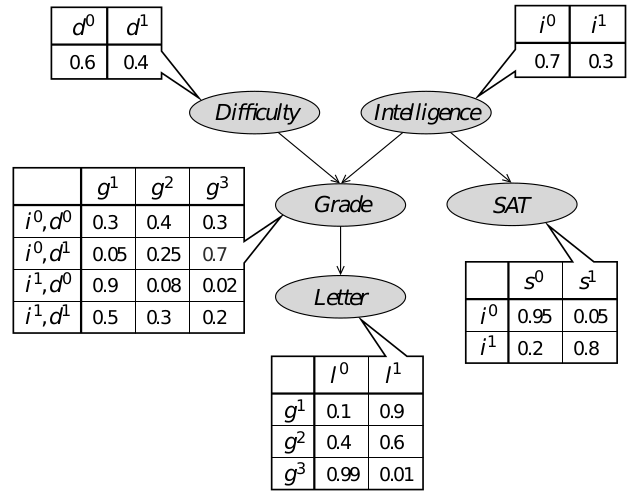}
\end{center}
\caption{Picture of the student Bayesian network, copied
  from~\cite{KollerF09}, with conditional probability tables.}
\label{fig:student}
\end{figure}
The graph of the student network is described in original form in
Figure~\ref{fig:student}. We see how a student's grade depends on the
difficulty of a test and the student's intelligence. The SAT score
only depends on intelligence; whether or not the student gets a strong
($l^1$) or weak ($l^0$) recommendation letter depends on the grade.

With each of the five nodes in the network a sample space is
associated, namely:
\[ D = \{d^{0}, d^{1}\}, \quad
I = \{i^{0}, i^{1}\}, \quad
G = \{g^{1}, g^{2}, g^{3}\}, \quad
S = \{s^{0}, s^{1}\}, \quad
L = \{l^{0}, l^{1}\}. \]

\noindent For the two inital nodes Difficulty (D) and Intelligence (I)
we obtain two distributions/states $\omega_{D}$ and $\omega_{I}$,
whose probabilities are given in the two upper tables in
Figure~\ref{fig:student}:
\[ \begin{array}{rclcrcl}
\omega_{D}
& = &
0.6\ket{d^0} + 0.4\ket{d^1}
& \mbox{\qquad} &
\omega_{I}
& = &
0.7\ket{i^0} + 0.3\ket{i^1}.
\end{array} \]

\noindent They capture the a priori state of affairs, with a $0.4$
likelihood of a difficult test ($d^1$), and a $0.3$ likelihood of an
intelligent student ($i^1$).

The remaining three nodes Grade (G), Letter (L) and SAT (S) have
incoming arrows from parent nodes, and are thus not initial. They
correspond to three channels:
\[ c_{G}\colon\kto{D\times I}{G}, \qquad
c_{S} \colon \kto{I}{S}, \qquad
c_{L} \colon \kto{G}{L}. \]

\noindent The definitions of these channels can be read directly 
from the tables. The SAT channel $c_{S} \colon \kto{I}{S}$ 
and the Letter channel $c_{L} \colon \kto{G}{L}$ are thus of the form:
\[ \begin{array}{ccc}
\begin{array}{rcl}
c_{S}(i^{0}) 
& = &
0.95\ket{s^0} + 0.05\ket{s^1}
\\
c_{S}(i^{1}) 
& = &
0.2\ket{s^0} + 0.8\ket{s^1}
\end{array}
& \mbox{\hspace*{5em}} &
\begin{array}{rcl}
c_{L}(g^{1}) 
& = &
0.1\ket{l^0} + 0.9\ket{l^1}
\\
c_{L}(g^{2}) 
& = &
0.4\ket{l^0} + 0.6\ket{l^1}
\\
c_{L}(g^{3}) 
& = &
0.99\ket{l^0} + 0.01\ket{l^1}
\end{array}
\end{array} \]

\noindent The Grade channel $c_{G}\colon\kto{D\times I}{G}$ looks as
follows.
\[ \begin{array}{rclcrcl}
c_{G}(d^{0}, i^{0}) 
& = &
0.3\ket{g^1} + 0.4\ket{g^2} + 0.3\ket{g^3}
& \mbox{\quad} &
c_{G}(d^{0}, i^{1}) 
& = &
0.9\ket{g^1} + 0.08\ket{g^2} + 0.02\ket{g^3}
\\
c_{G}(d^{1}, i^{0}) 
& = &
0.05\ket{g^1} + 0.25\ket{g^2} + 0.7\ket{g^3}
& &
c_{G}(d^{1}, i^{1}) 
& = &
0.5\ket{g^1} + 0.3\ket{g^2} + 0.2\ket{g^3}
\end{array} \]

\noindent (Notice that we switched the order of $i$ and $d$ wrt.\ the
tables in Figure~\ref{fig:student}; we have done so in order to remain
consistent with the order of the inputs $D$ and $I$ as suggested in
the network in Figure~\ref{fig:student}. This is actually a subtle
issue, because usually in graphs there is no order on the parents of a
node, that is, the parents form a \emph{set} and not a \emph{list}.)

We now discuss a number of inference questions from~\cite{KollerF09}
and illustrate how they are answered systematically using our
perspective with states, predicates and channels.
\begin{enumerate}
\item \label{question:apriori} \textbf{What are the a priori probabilities
  for the recommendation?}  To answer this question we follow the
  graph in Figure~\ref{fig:student} and see that the answer is given
  by twice using state transformation, namely:
\[ \begin{array}{rcl}
c_{L} \gg \big(c_{G} \gg (\omega_{D}\otimes\omega_{I})\big)
& = &
0.498\ket{l^0} + 0.502\ket{l^1}, \mbox{\qquad or, equivalently,}
\\
& = &
\big(c_{L} \klafter c_{G}\big) \gg (\omega_{D}\otimes\omega_{I}).
\end{array} \]

\item \label{question:intel} \textbf{What if we know that the student is
  not intelligent?}  The non-intelligence translates into the point
  predicate $\indic{\{i^0\}}$ on the set $I$, which we use to update
  the intelligence state $\omega_{I}$ before doing the same state
  transformations:
\[ \begin{array}{rcl}
c_{L} \gg \big(c_{G} \gg (\omega_{D}\otimes (\omega_{I}|_{\indic{\{i^0\}}}))\big)
& = &
0.611\ket{l^0} + 0.389\ket{l^1}.
\end{array} \]

\item \label{question:easy} \textbf{What if we additionally know that
  the test is easy?}  The easiness evidence translates into the
  predicate $\indic{\{d^0\}}$ on $D$, which is used for updating the
  difficulty state:
\[ \begin{array}{rcl}
c_{L} \gg \big(c_{G} \gg 
   ((\omega_{D}|_{\indic{\{d^0\}}})\otimes (\omega_{I}|_{\indic{\{i^0\}}}))\big)
& = &
0.487\ket{l^0} + 0.513\ket{l^1} 
\\
& = &
\big(c_{L} \klafter c_{G}\big) \gg
   \big((\omega_{D} \otimes \omega_{I})|_{(\indic{\{d^0\}}\otimes \indic{\{i^0\}})}\big).
\end{array} \]

\noindent The previous two outcomes are obtained by what is called
`causal reasoning' or `prediction' or `forward inference', see the
table at the end of Subsection~\ref{subsec:transformation}. We
continue with `backward inference', also called `evidential reasoning'
or `explanation'.

\medskip

\item \label{question:cgrade} \textbf{What is the intelligence given a
  C-grade ($g^3$)?} The evidence predicate $\indic{\{g^3\}}$ is a
  predicate on the set $G$.  We like to learn about the revised
  intelligence. This is done as follows.  Via predicate transformation
  we obtain a predicate $c_{G} \ll \indic{\{g^3\}}$ on $D\times I$. We
  can use it to update the product state
  $\omega_{D}\otimes\omega_{I}$. We then get the update intelligence
  by taking the second marginal, as in:
\[ \begin{array}{rcl}
\sM\big((\omega_{D}\otimes\omega_{I})\big|_{c_{G} \ll \indic{\{g^3\}}}\big)
& = &
0.921\ket{i^0} + 0.0789\ket{i^1}.
\end{array} \]

\noindent We see that the new intelligence ($i^1$) is significantly
lower than the a priori value of $0.3$, once a low grade is
observed. The updated difficulty ($d^1$) probability is higher than
the original $0.4$; it is obtained by taking the first marginal:
\[ \begin{array}{rcl}
\fM\big((\omega_{D}\otimes\omega_{I})\big|_{c_{G} \ll \indic{\{g^3\}}}\big)
& = &
0.371\ket{d^0} + 0.629\ket{d^1}.
\end{array} \]

\item \label{question:weak} \textbf{What is the intelligence given a
  weak recommendation?}  We now have a point predicate
  $\indic{\{l^0\}}$ on the set $L$. Hence we have to do predicate
  transformation twice, along the channels $c_{L}$ and $c_{G}$, in
  order to reach the initial states. This is done as:
\[ \begin{array}{rcl}
\sM\big((\omega_{D}\otimes\omega_{I})\big|_{c_{G} \ll (c_{L} \ll \indic{\{l^0\}})}\big)
& = &
0.86\ket{i^0} + 0.14\ket{i^1}, \mbox{\qquad or, equivalently,}
\\
& = &
\sM\big((\omega_{D}\otimes\omega_{I})\big|_{(c_{L} \klafter c_{G}) \ll 
   \indic{\{l^0\}}}\big).
\end{array} \]

\item \textbf{What is the intelligence given a C-grade but a high SAT
  score?} We now have two forms of evidence, namely the point
  predicate $\indic{\{g^3\}}$ on $G$ for the C-grade, and the point
  predicate $\indic{\{s^1\}}$ on $S$ for the high SAT score.  We can
  transform the latter to a predicate $c_{S} \ll \indic{\{s^1\}}$ on
  the set $I$ and update the state $\omega_{I}$ with it. Then we can
  procede as in question~\ref{question:cgrade}:
\[ \begin{array}{rcl}
\sM\big((\omega_{D}\otimes(\omega_{I}|_{c_{S} \ll \indic{\{s^1\}}}))
   \big|_{c_{G} \ll \indic{\{g^3\}}}\big)
& = &
0.422\ket{i^0} + 0.578\ket{i^1}.
\end{array} \]

\noindent Thus the probability of high intelligence is $57.8\%$ under
these circumstances.

Using calculation rule that we have seen before, see notably in
Lemma~\ref{lem:conditioning}, this intelligence distribution can also
be computed by weakening the predicate $c_{S} \ll \indic{\{s^1\}}$ to
$\sW(c_{S} \ll \indic{\{s^1\}})$ on $D\times I$.  Then we can take the
conjunction with $c_{G} \ll \indic{\{g^3\}}$ and perform a single
update, as in:
\[ \sM\big((\omega_{D}\otimes\omega_{I})
   \big|_{\subsW(c_{S} \ll \indic{\{s^1\}}) \andthen (c_{G} \ll \indic{\{g^3\}})}\big) \]

\noindent But one can also do the update with $c_{S} \ll \indic{\{s^1\}}$
at the very end, after the marginalisation, as in:
\[ \sM\big((\omega_{D}\otimes\omega_{I})
   \big|_{c_{G} \ll \indic{\{g^3\}}}\big)\big|_{c_{S} \ll \indic{\{s^1\}}} \]

\noindent The associated difficulty level is the first marginal:
\[ \begin{array}{rcl}
\fM\big((\omega_{D}\otimes(\omega_{I}|_{\indic{\{s^1\}}}))
   \big|_{c_{G} \ll \indic{\{g^3\}}}\big)
& = &
0.24\ket{d^0} + 0.76\ket{d^1}.
\end{array} \]
\end{enumerate}

\noindent The answers to the above questions hopefully convey the
systematic thinking that is behind the use of channels --- in forward
or backward manner, following the network structure --- in order to
capture the essence of Bayesian networks. This systematics is
elaborated further in subsequent sections. In the above `student'
example we have obtained the same outcomes as in traditional
approaches. We conclude with an illustration where things differ.

\begin{example}
\label{ex:burglar}
The power of the channel-based approach is that it provides a `logic'
for Bayesian inference, giving high-level expressions $c \gg
\omega|_{p}$ and $\omega|_{c \ll q}$ for forward and backward
inference. We include an illustration from~\cite{Barber12} where our
method produces a different outcome. The logical description may help
to clarify the differences.

Consider the following Bayesian network.
\[ \hspace*{3em}\xymatrix@C-1.5pc@R-1.5pc{
{\setlength\tabcolsep{0.2em}
   \renewcommand{\arraystretch}{1}
\begin{tabular}{|c|}
\hline
$\Prob$(burglar) \\
\hline\hline
$0.01$ \\
\hline
\end{tabular}}
& \ovalbox{\strut burglar}\ar[ddr] & & 
{\hspace*{-1em}\ovalbox{\strut earthquake}\hspace*{-1em}}\ar[ddl]\ar[ddr]
   \rlap{\qquad\smash{\setlength\tabcolsep{0.2em}\renewcommand{\arraystretch}{1}
\begin{tabular}{|c|}
\hline
$\Prob$(earthquake) \\
\hline\hline
$0.000001$ \\
\hline
\end{tabular}}} 
\\
\\
& & \ovalbox{\strut alarm}
  \llap{\smash{\setlength\tabcolsep{0.2em}\renewcommand{\arraystretch}{1}
\begin{tabular}{|c|c|c|}
\hline
burglar & earthquake & $\Prob$(alarm) \\
\hline\hline
$b$ & $e$ & $0.9999$ \\
\hline
$b$ & $\no{e}$ & $0.99$ \\
\hline
$\no{b}$ & $e$ & $0.99$ \\
\hline
$\no{b}$ & $\no{e}$ & $0.0001$ \\
\hline
\end{tabular}}\hspace*{5em}}
& &
\ovalbox{\strut radio} &
{\setlength\tabcolsep{0.2em}\renewcommand{\arraystretch}{1}
\begin{tabular}{|c|c|}
\hline
earthquake & $\Prob$(radio) \\
\hline\hline
$e$ & $1$ \\
\hline
$\no{e}$ & $0$ \\
\hline
\end{tabular}}
}\]

\medskip

\noindent In this case we have binary sets $B = \{b, \no{b}\}$, $E =
\{e, \no{e}\}$, $A = \{a, \no{a}\}$ and $R = \{r, \no{r}\}$ with
initial states $\omega_{B} = 0.01\ket{b} + 0.99\ket{\no{b}}$ and
$\omega_{E} = 0.000001\ket{e} + 0.999999\ket{\no{e}}$. There are two
channels $c_{A} \colon \kto{B\times E}{A}$ and $c_{R} \colon
\kto{E}{R}$ based on the above (two lower) tables.

The following questions are asked in~\cite[Example~3.1
  and~3.2]{Barber12}.
\begin{enumerate}
\item \label{question:alarm} \textbf{What is the probability of a
  burglary given that the alarm sounds?} In this case we have evidence
  $\indic{\{a\}}$ on the set $A$, we pull it back to $B\times E$ along
  the channel $c_A$, and we update the joint state
  $\omega_{B}\otimes\omega_{E}$ and take the first marginal:
\[ \begin{array}{rcl}
\fM\big((\omega_{B}\otimes\omega_{E})\big|_{c_{A} \ll \indic{\{a\}}}\big)
& = &
0.99000198\ket{b} + 0.00999802\ket{\no{b}}.
\end{array} \]

\item \label{question:radio} \textbf{What is this probability if we
  additionally hear a warning on the radio?} In that case we have
  additional evidence $\indic{\{r\}}$ on $R$, which is pulled back
  along the channel $c_R$ and used to update the state $\omega_{E}$.
  Then:
\[ \begin{array}{rcl}
\fM\big((\omega_{B}\otimes(\omega_{E}|_{c_{R} \ll \indic{\{r\}}}))\big|_{c_{A} \ll \indic{\{a\}}}\big)
& = &
0.010099\ket{b} + 0.989901\ket{\no{b}}.
\end{array} \]

\item \label{question:soft} \ldots \textbf{``imagine that we are only
  70\% sure we heard the burglar alarm sounding''} In this situation
  we have a fuzzy predicate $q \colon A \rightarrow [0,1]$ with $q(a)
  = 0.7$ and $q(\no{a}) = 0.3$.  We perform the same computation as in
  question~\ref{question:alarm}, but now with evidence $q$ instead of
  $\indic{\{a\}}$. This yields:
\begin{equation}
\label{eqn:softpullback}
\begin{array}{rcl}
\fM\big((\omega_{B}\otimes\omega_{E})\big|_{c_{A} \ll q}\big)
& = &
0.0229\ket{b} + 0.9771\ket{\no{b}}.
\end{array}
\end{equation}

\noindent However, in~\cite{Barber12} a completely different
computation is performed. The assumption about the alarm is not
interpreted as a predicate, but as a state $\sigma = 0.7\ket{a} +
0.3\ket{\no{a}}$, even though the whole example is presented as an
illustration of the use of \emph{soft} evidence. A different predicate
$p\colon A \rightarrow [0,1]$ is constructed, namely:
\begin{equation}
\label{eqn:softpred}
\begin{array}{rcl}
p(x)
& = &
\fM\big((\omega_{B}\otimes\omega_{E})\big|_{c_{A} \ll \indic{\{x\}}}\big)
  \models \indic{\{b\}}.
\end{array}
\end{equation}

\noindent Thus, $p(x)$ is the probability of a burglary if the alarm
is $x$ (that is, ``sounding'' if $x = a$ and ``silent'' if $x =
\no{a}$). The answer to the question ``What is the probability of a
burglary under this soft-evidence?'' in~\cite{Barber12} is computed
as:
\begin{equation}
\label{eqn:softstate}
\begin{array}{rcccl}
\sigma\models p
& \,=\, &
0.7\cdot p(a) + 0.3\cdot p(\no{a})
& = &
0.69303.
\end{array}
\end{equation}
\end{enumerate}

\noindent For questions~\ref{question:alarm} and~\ref{question:radio}
our calculations coincide with the ones in~\cite{Barber12}, but for
question~\ref{question:soft} the answers clearly differ. We briefly
analyse the situation.
\begin{itemize}
\item The description of \emph{soft} evidence in~\cite{Barber12}
  says\footnote{Please keep the difference in terminology in mind: a
    `state' in~\cite{Barber12} is what we would call an element of the
    sample space; it is not a distribution as used here.}: ``In soft
  or uncertain evidence, the evidence variable is in more than one
  state, with the strength of our belief about each state being given
  by probabilities.'' In subsequent illustrations these probabilities
  add up to $1$. This strongly suggests that soft evidence
  in~\cite{Barber12} is not evidence in the form of a predicate, but
  is a distribution (state). Indeed, the
  computation~\eqref{eqn:softstate} uses the given soft evidence as a
  state $\sigma$. Here we see a clear case of mixing up states and
  predicates, \emph{c.f.}~Subsection~\ref{subsec:distinguish}.

\item In contrast, in the current setting a \emph{fuzzy} predicate is
  a $[0,1]$-valued function, without any requirement that
  probabilities add up to $1$. Hence in the setting of the above
  example we could have a fuzzy predicate saying: we are 70\% sure we
  heard the alarm and 20\% sure that we heard no alarm. This would
  translate into $q(a) = 0.7$ and $q(\no{a}) = 0.2$. In that case we
  would still be able to do the computation~\eqref{eqn:softpullback},
  but the approach of~\cite{Barber12} would fail.


\item Apart from the state/predicate confusion, the difference in
  computation can be formulated as: at which stage of the computation
  does one need to weigh the softness of the evidence? At the very
  end, as in~\eqref{eqn:softstate} following~\cite{Barber12}, or right
  at the beginning, as in~\eqref{eqn:softpullback}. The (soft)
  evidence about the alarm is translated in~\eqref{eqn:softpred}, via
  the predicate $p$, into (soft) evidence about the burglary, which is
  then weighed in~\eqref{eqn:softstate} using the weights $0.7$ and
  $0.3$ that were originally given about the alarm (and not about the
  burglary).

We claim that the weighing should be done in the beginning, on the
alarm data for which the original evidence was given. In this way the
channel $c_{A} \colon \kto{B\times E}{A}$ takes the different alarm
evidence probabilities into account, and translates them, via
predicate transformation, into (soft) evidence on burglary and
earthquake. More precisely, recall that we formalise the original soft
evidence $q$ on $A$ as $q(a) = 0.7, q(\no{a}) = 0.3$.  What evidence
does this give us on $B\times E$? The only reasonable answer is the
transformed predicate $c_{A} \ll q \colon B\times E \rightarrow
[0,1]$, with outcomes:
\[ \begin{array}{rcl}
\big(c_{A} \ll q\big)(b,e)
& = &
\sum_{x\in A} c_{A}(b,e)(x) \cdot q(x)
\hspace*{\arraycolsep}=\hspace*{\arraycolsep}
0.9999\cdot 0.7 + 0.0001\cdot 0.3
\hspace*{\arraycolsep}=\hspace*{\arraycolsep}
0.7
\end{array} \]

\noindent Similarly one computes:
\[ \begin{array}{rccclcrcl}
\big(c_{A} \ll q\big)(b,\no{e})
& = &
\big(c_{A} \ll q\big)(\no{b},e)
& = &
0.696
& \hspace*{4em} &
\big(c_{A} \ll q\big)(\no{b},\no{e})
& = &
0.3.
\end{array} \]

\noindent The last equation expresses: the alarm evidence $q$ gives me
evidence that there was not a burglary and also not an earthquake with
$30\%$ certainty. There are similar interpretations of the other three
cases in $B\times E$. With this transformed evidence $c_{A} \ll q$ on
$B\times E$ we can update the product joint state
$\omega_{B}\otimes\omega_{E}$ on $B\times E$. It yields the state
$(\omega_{B}\otimes\omega_{E})|_{c_{A} \ll q}$, whose first marginal
yields the required burglary probability, as computed
in~\eqref{eqn:softpullback}.

\item Yet another perspective is that the above computation and the
  one in~\cite{Barber12} are based on different ways of understanding
  what soft evidence actually means. In~\cite{Barber12} this notion,
  even though it is not made mathematically precise, appears to have
  an ontological interpretation: ``the alarm was heard'' is a new
  fact, which has 70\% chances of being true, and is therefor used as
  a state (distribution). On the other hand, our fuzzy predicate
  interpretation has an epistemological flavour: it is new
  information about a (possibly inconsistent) agent's belief that is
  made available. For instance, it can take the form of the testimony
  of a confused (or drunken) witness, saying: yes, I'm absolutely sure
  I heard the alarm; and also, when asked next, the witness could say:
  I'm certain I heard no alarm. We would then have soft evidence $p$
  with $p(a) = p(\no{a}) = 1$, and thus a fuzzy predicate rather than
  a probability distribution.

\end{itemize}

\noindent We shall briefly return to these different was of
computation in Example~\ref{ex:burglarjoint} where we show that the
outcome in~\eqref{eqn:softpullback} also appears via `crossover
inference'.

\end{example}

\section{String diagrams for Bayesian probability}\label{sec:string}

Abstractly, channels are arrows of a category, which is
\emph{symmetric monoidal}: it has sequential $\klafter$ and parallel
$\otimes$ composition. This categorical structure enables the use of a
graphical (yet completely formal) notation for channels in terms of
\emph{string diagrams}~\cite{Selinger2009}. We have no intention of
giving a complete account of the string diagrammatic calculus
here, and refer instead to~\cite{Fong12} and~\cite[Remark 3.1]{JacobsZ16} for details. Nonetheless, it is worthwhile pointing similarities and differences
between the graphical representation of channels as string diagrams
and the usual Bayesian network notation. We shall also use string
diagrams to give a pictorial account of the important notion of
disintegration (in the next section).

Informally speaking, string diagrams for channels are similar to the
kind of graphs that is used for Bayesian networks, see
Figure~\ref{fig:student}, but there are important differences.
\begin{enumerate}
\item Whereas flow in Bayesian networks is top-down, we will adopt the
  convention that in string diagrams the flow is bottom-up. This is an
  non-essential, but useful difference, because it makes immediately
  clear in the current context whether we are dealing with a Bayesian
  network or with a string diagram. Also, it makes our presentation
  uniform with previous work, see \textit{e.g.}~\cite{ChoJ17a}.

\item The category where channels are arrows has extra structure,
  which allows for the use of ``special'' string diagrams representing
  certain elementary operations. We will have explicit string diagrams
  for \emph{copying} and \emph{discarding} variables, namely:
\[ \mbox{copy}
\quad
=
\quad
\vcenter{\hbox{%
\begin{tikzpicture}[font=\small]
\node[copier] (c) at (0,0) {};
\coordinate (x1) at (-0.3,0.3);
\coordinate (x2) at (0.3,0.3);
\draw (c) to[out=165,in=-90] (x1);
\draw (c) to[out=15,in=-90] (x2);
\draw (c) to (0,-0.3);
\end{tikzpicture}}}
\mbox{\qquad\qquad and \qquad\qquad}
\mbox{discard}
\quad
=
\quad
\vcenter{\hbox{%
\begin{tikzpicture}[font=\small]
\node[discarder] (d) at (0,0) {};
\draw (d) to (0,-0.3);
\end{tikzpicture}}}
\]

\noindent There are some `obvious' equations between diagrams involving
such copy and discard, such as:
\[
\vcenter{\hbox{%
\begin{tikzpicture}[font=\small]
\node[copier] (c) at (0,0) {};
\coordinate (x1) at (-0.3,0.3);
\node[discarder] (d) at (x1) {};
\coordinate (x2) at (0.3,0.5);
\draw (c) to[out=165,in=-90] (x1);
\draw (c) to[out=15,in=-90] (x2);
\draw (c) to (0,-0.3);
\end{tikzpicture}}}
\;=\;
\vcenter{\hbox{%
\begin{tikzpicture}[font=\small]
\draw (0,0) to (0,0.8);
\end{tikzpicture}}}
\;=\;
\vcenter{\hbox{%
\begin{tikzpicture}[font=\small]
\node[copier] (c) at (0,0) {};
\coordinate (x1) at (-0.3,0.5);
\coordinate (x2) at (0.3,0.3);
\node[discarder] (d) at (x2) {};
\draw (c) to[out=165,in=-90] (x1);
\draw (c) to[out=15,in=-90] (x2);
\draw (c) to (0,-0.3);
\end{tikzpicture}}}
\qquad\quad
\vcenter{\hbox{%
\begin{tikzpicture}[font=\small]
\node[copier] (c2) at (0,0) {};
\node[copier] (c1) at (-0.3,0.3) {};
\draw (c2) to[out=165,in=-90] (c1);
\draw (c2) to[out=15,in=-90] (0.4,0.6);
\draw (c1) to[out=165,in=-90] (-0.6,0.6);
\draw (c1) to[out=15,in=-90] (0,0.6);
\draw (c2) to (0,-0.3);
\end{tikzpicture}}}
\;=\;
\vcenter{\hbox{%
\begin{tikzpicture}[font=\small]
\node[copier] (c2) at (-0.5,0) {};
\node[copier] (c1) at (-0.2,0.3) {};
\draw (c2) to[out=15,in=-90] (c1);
\draw (c2) to[out=165,in=-90] (-1,0.6);
\draw (c1) to[out=165,in=-90] (-0.5,0.6);
\draw (c1) to[out=15,in=-90] (0.1,0.6);
\draw (c2) to (-0.5,-0.3);
\end{tikzpicture}}}
\qquad\quad
\vcenter{\hbox{%
\begin{tikzpicture}[font=\small]
\node[copier] (c) at (0,0.4) {};
\draw (c)
to[out=15,in=-90] (0.25,0.7)
to[out=90,in=-90] (-0.25,1.4);
\draw (c)
to[out=165,in=-90] (-0.25,0.7)
to[out=90,in=-90] (0.25,1.4);
\draw (c) to (0,0.1);
\end{tikzpicture}}}
\;=\;
\vcenter{\hbox{%
\begin{tikzpicture}[font=\small]
\node[copier] (c) at (0,0.4) {};
\draw (c)
to[out=15,in=-90] (0.25,0.7);
\draw (c)
to[out=165,in=-90] (-0.25,0.7);
\draw (c) to (0,0.1);
\end{tikzpicture}}}
\]
These equations represent the fact that copy is the multiplication and discard is the unit of a commutative monoid.
\item With string diagrams one can clearly express joint states, on
  product domains like $X_{1}\times X_{2}$, or $X_{1} \times \cdots
  \times X_{n}$. This is done by using multiple outgoing pins, coming
  out of a triangle shape --- used for states --- as for
  $\omega\in\Dst(X_{1}\times X_{2})$ and $\sigma\in\Dst(X_{1}\times
  \cdots \times X_{n})$ in:
\[ \vcenter{\hbox{%
\begin{tikzpicture}[font=\small]
\node[state] (omega) at (0,0) {\;$\omega$\;};
\coordinate (X) at (-0.25,0.5) {};
\coordinate (Y) at (0.25,0.5) {};
\draw (omega) ++(-0.25, 0) to (X);
\draw (omega) ++(0.25, 0) to (Y);
\node at (-0.55,0.4) {$X_1$};
\node at (0.55,0.4) {$X_2$};
\end{tikzpicture}}}
\hspace*{4em}
\vcenter{\hbox{%
\begin{tikzpicture}[font=\small]
\node[state] (omega) at (0,0) {$\;\;\sigma\;\;$};
\coordinate (X) at (-0.5,0.5) {};
\coordinate (Y) at (0.5,0.5) {};
\draw (omega) ++(-0.5, 0) to (X);
\draw (omega) ++(0.5, 0) to (Y);
\node at (-0.75,0.4) {$X_1$};
\node at (0.0,0.4) {$\cdots$};
\node at (0.75,0.4) {$X_n$};
\end{tikzpicture}}}
\]

\noindent With this notation in place we can graphically express the
marginals via discarding $\ground$ of wires:
\[ \margsign_{1}(\omega)
\quad
=
\quad\vcenter{\hbox{%
\begin{tikzpicture}[font=\small]
\node[state] (omega) at (0,0) {\;$\omega$\;};
\coordinate (X) at (-0.25,0.55) {};
\node [discarder] (Y) at (0.25,0.4) {};
\draw (omega) ++(-0.25, 0) to (X);
\draw (omega) ++(0.25, 0) to (Y);
\node at (-0.55,0.4) {$X_1$};
\end{tikzpicture}}}
\mbox{\qquad\qquad and \qquad\qquad}
\margsign_{2}(\omega)
\quad
=
\quad\vcenter{\hbox{%
\begin{tikzpicture}[font=\small]
\node[state] (omega) at (0,0) {\;$\omega$\;};
\node [discarder] (X) at (-0.25,0.4) {};
\coordinate (Y) at (0.25,0.55) {};
\draw (omega) ++(-0.25, 0) to (X);
\draw (omega) ++(0.25, 0) to (Y);
\node at (0.55,0.4) {$X_2$};
\end{tikzpicture}}}
\]

\item Channels are \emph{causal} or \emph{unitary} in the sense that
  discarding their output is the same as discarding their input:
\[
\vcenter{\hbox{%
\begin{tikzpicture}[font=\small]
\node[arrow box] (c) at (0,0) {$c$};
\node[discarder] (d) at (0,0.5) {};
\draw  (c) to (d);
\draw (c) to (0,-0.5);
\end{tikzpicture}}}
\;=\;
\vcenter{\hbox{%
\begin{tikzpicture}[font=\small]
\node[discarder] (d) at (0,0) {};
\draw (d) to (0,-0.5);
\end{tikzpicture}}}
\]
\end{enumerate}

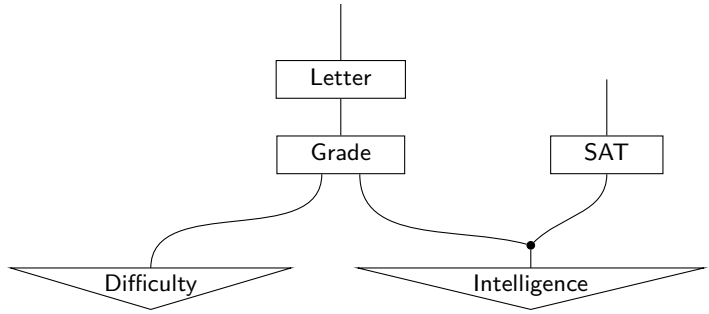
\begin{figure}
\begin{center}
$\vcenter{\hbox{%
\begin{tikzpicture}[font=\small]
\node[copier] (copier) at (2.5,0.3) {};
\node[state] (dif) at (-2.5,0) {\textsf{Difficulty}};
\node[state] (int) at (2.5,0) {\textsf{Intelligence}};
\node[arrow box] (sat) at (3.5,1.5) {\hspace*{1em}\textsf{SAT}\hspace*{1em}};
\node[arrow box] (gra) at (0,1.5) {\hspace*{1em}\textsf{Grade}\hspace*{1em}};
\node[arrow box] (let) at (0,2.5) {\hspace*{1em}\textsf{Letter}\hspace*{1em}};
\coordinate (Y) at (-0.1,1.2);
\draw (dif) to[out=90,in=-90] ([xshiftu=-0.25]gra.south);
\draw (int) to (copier);
\draw (copier) to[out=165,in=-90] ([xshiftu=0.25]gra.south);
\draw (copier) to[out=45,in=-90] (sat);
\draw (sat) to ([yshiftu=1]sat);
\draw (gra) to (let);
\draw (let) to ([yshiftu=1]let);
\end{tikzpicture}}}$
\end{center}
\caption{Student network from Figure~\ref{fig:student} expressed as
  string diagram.}
\label{fig:studentstring}
\end{figure}

The Intelligence node in Figure~\ref{fig:student} has two outgoing
arrows, but this does not mean that Intelligence is a joint
state. Instead, these two arrows indicate that the outgoing wire
should be copied, with one copy going to the Grade node and one to the
SAT node. In string diagram notation this copying is written
explicitly as in the string-diagrammatic analogue of the student
network in Figure~\ref{fig:studentstring}.

Recall that we wrote $\omega_{D} = 0.6\ket{d^0} + 0.4\ket{d^1}$ and
$\omega_{I} = 0.7\ket{i^0} + 0.3\ket{i^1}$ for the initial states of
the student network. The product state
\[ \begin{array}{rcl}
\omega_{D}\otimes\omega_{I}
& = &
0.42\ket{d^{0},i^{0}} + 0.18\ket{d^{0},i^{1}} + 
   0.28\ket{d^{1},i^{0}} + 0.12\ket{d^{1},i^{1}}
\end{array} \]

\noindent is non-entwined, since it equals the product of its
marginals $\omega_{D}$ and $\omega_{I}$. A basic fact in probability
is that conditioning can create entwinedness, see
\textit{e.g.}~\cite{JacobsZ17} for more information. We can see this
concretely when the above product state $\omega_{D}\otimes\omega_{I}$
is conditioned as in the fourth question in the previous section:
\[ \begin{array}{rcl}
(\omega_{D}\otimes\omega_{I})\big|_{c_{G} \ll \indic{\{g^3\}}}
& = &
\frac{12600}{34636}\ket{d^{0},i^{0}} + \frac{36}{34636}\ket{d^{0},i^{1}}
  + \frac{19600}{34636}\ket{d^{1},i^{0}} + \frac{2400}{34636}\ket{d^{1},i^{1}}.
\end{array} \]

\auxproof{
We have:
\[ \begin{array}{rcl}
\omega_{D}\otimes\omega_{I}
& = &
0.42\ket{d^{0},i^{0}} + 0.18\ket{d^{0},i^{1}} + 
   0.28\ket{d^{1},i^{0}} + 0.12\ket{d^{1},i^{1}}
\\
(c_{G} \ll \indic{\{g^3\}})(d^{0},i^{0})
& = & 
0.3
\\
(c_{G} \ll \indic{\{g^3\}})(d^{0},i^{1})
& = & 
0.02
\\
(c_{G} \ll \indic{\{g^3\}})(d^{1},i^{0})
& = & 
0.7
\\
(c_{G} \ll \indic{\{g^3\}})(d^{1},i^{1})
& = & 
0.2
\\
\omega_{D}\otimes\omega_{I} \models c_{G} \ll \indic{\{g^3\}}
& = &
0.42\cdot 0.3 + 0.18 \cdot 0.002 + 0.28\cdot 0.7 + 0.12\cdot 0.2
\\
& = &
0.34636
\\
\omega_{D}\otimes\omega_{I}|_{c_{G} \ll \indic{\{g^3\}}}
& = &
\frac{0.126}{0.34636}\ket{d^{0},i^{0}} + \frac{0.00036}{0.34636}\ket{d^{0},i^{1}}
  + \frac{0.196}{0.34636}\ket{d^{1},i^{0}} + \frac{0.024}{0.34636}\ket{d^{1},i^{1}}
\\
& = &
\frac{12600}{34636}\ket{d^{0},i^{0}} + \frac{36}{34636}\ket{d^{0},i^{1}}
  + \frac{19600}{34636}\ket{d^{1},i^{0}} + \frac{2400}{34636}\ket{d^{1},i^{1}}.
\end{array} \]
}

\noindent With some effort one can show that this state is \emph{not}
the product of its marginals: it is entwined. In the language of
string diagrams we can express this difference by writing:
\[ \omega_{D}\otimes\omega_{I}
\quad
=
\quad
\vcenter{\hbox{%
\begin{tikzpicture}[font=\small]
\node[state] (s1) at (0,0) {$\;\;$};
\node[state] (s2) at (1,0) {$\;\;$};
\draw (s1) to (0, 0.5);
\draw (s2) to (1, 0.5);
\end{tikzpicture}}}
\hspace*{5em}
(\omega_{D}\otimes\omega_{I})\big|_{c_{G} \ll \indic{\{g^3\}}}
\quad
=
\quad
\vcenter{\hbox{%
\begin{tikzpicture}[font=\small]
\node[state] (s) at (0,0) {$\qquad$};
\draw (s) ++(-0.5, 0) to (-0.5, 0.5);
\draw (s) ++(0.5, 0) to (0.5, 0.5);
\end{tikzpicture}}}
\]

\section{From joint states to Bayesian networks}\label{sec:essence} 

Our framework allows to express states/distributions and Bayesian
networks as entities of the same kind, namely as channels. It is
natural to ask how the process of forming a Bayesian network from a
distribution can be integrated in the picture.

In traditional probability theory, this procedure forms one of the
original motivations for developing the notion of Bayesian network in
the first place. Such networks allow for more efficient representation
of probabilistic information (via probability tables, as in
Figure~\ref{fig:student}) than joint states, which quickly become
unmanageable via an exponential explosion. We quote~\cite{KollerF09}:
``\ldots the explicit representation of the joint distribution is
unmanageable from every perspective. Computationally, it is very
expensive to manipulate and generally too large to store in memory''
and~\cite{RusselN03}: ``\ldots a Bayesian network can often be far
more \emph{compact} than the full joint distribution''.

The procedure of forming a Bayesian network from a given state usually
goes through a sub-routine called \emph{disintegration}. For a
channel-based definition of disintegration, suppose we have a state
$\omega\in\Dst(X)$ and a channel $c\colon \kto{X}{Y}$. Then we can
form a joint state $\sigma\in\Dst(X\times Y)$ as described by the
following string diagram:
\begin{equation}
\label{diag:disintegration}
\vcenter{\hbox{%
\begin{tikzpicture}[font=\small]
\node[state] (s) at (0,0) {$\;\sigma\;$};
\draw (s) ++(-0.25, 0) to (-0.25, 0.4);
\draw (s) ++(0.25, 0) to (0.25, 0.4);
\end{tikzpicture}}}
\quad 
\coloneqq 
\quad
\vcenter{\hbox{%
\begin{tikzpicture}[font=\small]
\node[state] (omega) at (0,0) {$\omega$};
\node[copier] (copier) at (0,0.2) {};
\node[arrow box] (c) at (0.35,0.8) {$c$};
\coordinate (X) at (-0.35,1.3) {} {};
\coordinate (Y) at (0.35,1.3) {} {};
\draw (omega) to (copier);
\draw (copier) to[out=165,in=-90] (-0.35,0.55) to (X);
\draw (copier) to[out=15,in=-90] (c);
\draw (c) to (Y);
\end{tikzpicture}}}
\mbox{\qquad\quad that is \qquad}
\begin{array}{rcl}
\sigma(x,y)
& = &
\omega(x)\cdot c(x)(y).
\end{array}
\end{equation}

\noindent The state $\omega$ is determined as the first marginal
$\omega=\margsign_{1}(\sigma)$ of $\sigma$. This can be seen by
discarding $\ground$ the second wire --- on the left and on the right
in the above equation --- and using that channels are causal, and that
discarding one wire of a copy is the identity wire.

Disintegration is the process in the other direction, from
a joint state to a channel.

\begin{definition}
\label{def:disintegration}
Let $\sigma\in\Dst(X\times Y)$ be a joint state. A
\emph{disintegration} of $\sigma$ is a channel $c\colon \kto{X}{Y}$
for which the equation~\eqref{diag:disintegration} holds, where
$\omega = \margsign_{1}(\sigma)$.
\end{definition}

There is a standard formula for disintegration of a state $\sigma
\in\Dst(X\times Y)$, namely:
\begin{equation}
\label{eqn:disintegration}
\begin{array}{rcccl}
c(x)
& \coloneqq &
\margsign_{2}\big(\sigma\big|_{\indic{\{x\}}\otimes\one}\big)
& = &
\displaystyle\sum_{y} \frac{\sigma(x,y)}{\margsign_{1}(\sigma)(x)}\bigket{y}.
\end{array}
\end{equation}

\noindent We shall say that the channel $c$ is `extracted' from
$\sigma$, or also that $\sigma$ `factorises' via $c$ as
in~\eqref{diag:disintegration}. Intuitively, channel $c$ captures the
conditional probabilities expressed in traditional notation as
$\Prob_{\sigma}(y \mid x)$ via a distribution on $Y$ indexed by
elements $x\in X$.

Definition \ref{def:disintegration} gives the basic form of
disintegration. There are several variations, which are explored
in~\cite{ChoJ17a} as part of a more abstract account of this
notion. For instance, by swapping the domains one can also extract a
channel $\kto{Y}{X}$, in the other direction. Also, if $\sigma$ is a
joint state on $n$ domains, there are in principle $2^n$ ways of
extracting a channel, depending on which pins are marginalised out,
and which (other) ones are reconstructed via the channel. For
instance, a disintegration of $\omega \in \Dst(X \times Y \times Z)$
can also be a channel $c \colon \kto{Z}{X \times Y}$. This example
suggests a digression on a channel-based definition of conditional
independence: $X$ and $Y$ are conditionally independent in $\omega$
given $Z$, written as $\cind{X}{Y}{Z}$, if any such disintegration $c$
for $\omega$ can actually be decomposed into channels $c_1 \colon
\kto{Z}{X}$ and $c_2 \colon \kto{Z}{Y}$. In string diagrams:
\begin{equation} 
\label{eq:condind}
\vcenter{\hbox{%
\begin{tikzpicture}[font=\small]
\node[state] (omega) at (0,0) {\;$\omega$\;};
\draw (omega) ++(-0.4, 0) to +(0,0.3);
\draw (omega) to +(0,0.3);
\draw (omega) ++(0.4, 0) to +(0,0.3);
\path[font=\normalsize,
execute at begin node=\everymath{\scriptstyle}]
node at (-0.4,0.5) {$X$}
node at (0,0.5) {$Y$}
node at (0.4,0.5) {$Z$};
\end{tikzpicture}}}
\;\;=\;\;
\vcenter{\hbox{%
\begin{tikzpicture}[font=\small]
\node[state] (omega) at (0,0) {\;$\omega_3$};
\node[copier] (c) at (0,0.2) {};
\node[arrow box] (a) at (-0.7,0.8) {\;\;\;\;\;c\;\;\;\;\;};
\draw (omega) to (c);
\draw (c) to[out=165,in=-90] (a);
\draw (-1,1.05) to (-1,1.3);
\draw (-0.3,1.05) to (-0.3,1.3);
\draw (c) to[out=15,in=-90] (0.7,0.6)
to (0.7,1.3);
\path[font=\normalsize,
execute at begin node=\everymath{\scriptstyle}]
node at (-1,1.5) {$X$}
node at (-0.3,1.5) {$Y$}
node at (0.7,1.5) {$Z$};
\end{tikzpicture}}}
\;\;=\;\;
\vcenter{\hbox{%
\begin{tikzpicture}[font=\small]
\node[state] (omega) at (0,0) {\;$\omega_3$};
\node[copier] (c) at (0,0.2) {};
\node[arrow box] (a) at (-0.7,0.8) {$c_1$};
\node[arrow box] (b) at (0,0.8) {$c_2$};
\draw (omega) to (c);
\draw (c) to[out=165,in=-90] (a);
\draw (a) to (-0.7,1.3);
\draw (c) to (b);
\draw (b) to (0,1.3);
\draw (c) to[out=15,in=-90] (0.7,0.6)
to (0.7,1.3);
\path[font=\normalsize,
execute at begin node=\everymath{\scriptstyle}]
node at (-0.7,1.5) {$X$}
node at (0,1.5) {$Y$}
node at (0.7,1.5) {$Z$};
\end{tikzpicture}}}
\end{equation}

\noindent where $\omega_{3} = \margsign_{3}(\omega) =
\Prob_{\omega}(z)$ is the third marginal. These channels $c_{1},
c_{2}$ may also be obtained by disintegration from the state
$\margsign_{1,2}(\omega) = \Prob_{\omega}(x,y)$ obtained by
marginalising out the third variable. In more traditional notation,
one can intuitively read \eqref{eq:condind} as saying that
$\Prob_{\omega}(x,y, z) = \Prob_{\omega}(x \mid z) \cdot
\Prob_{\omega}(y\mid z) \cdot \Prob_{\omega}(z)$. We refer
to~\cite{ChoJ17a} for the adequacy of this definition of conditional
independence and its properties.

\medskip

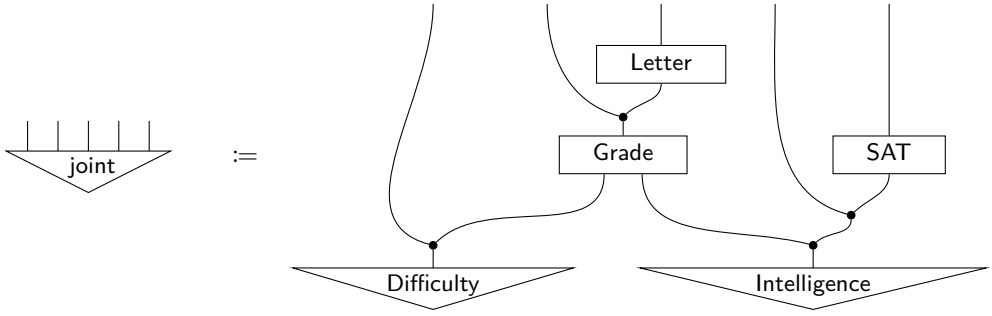
\begin{figure}
\begin{center}
$\vcenter{\hbox{%
\begin{tikzpicture}[font=\small]
\node[state] (joint) at (0,0) {\textsf{ joint }};
\draw (joint) ++(-0.8, 0) to (-0.8, 0.4);
\draw (joint) ++(-0.4, 0) to (-0.4, 0.4);
\draw (joint) ++(0.0, 0) to (0.0, 0.4);
\draw (joint) ++(0.4, 0) to (0.4, 0.4);
\draw (joint) ++(0.8, 0) to (0.8, 0.4);
\end{tikzpicture}}}
\qquad
\coloneqq
\quad
\vcenter{\hbox{%
\begin{tikzpicture}[font=\small]
\node[state] (dif) at (-2.5,0) {\textsf{Difficulty}};
\node[copier] (copierD) at (-2.5,0.3) {};
\node[state] (int) at (2.5,0) {\textsf{Intelligence}};
\node[copier] (copierI1) at (2.5,0.3) {};
\node[copier] (copierI2) at (3.0,0.7) {};
\node[arrow box] (sat) at (3.5,1.5) {\hspace*{1em}\textsf{SAT}\hspace*{1em}};
\node[arrow box] (gra) at (0,1.5) {\hspace*{1em}\textsf{Grade}\hspace*{1em}};
\node[copier] (copierG) at (0,2.0) {};
\node[arrow box] (let) at (0.5,2.7) {\hspace*{1em}\textsf{Letter}\hspace*{1em}};
\coordinate (Y) at (-0.1,1.2);
\draw (dif) to (copierD);
\draw (copierD) to[out=165,in=-90] (-2.5,3.5);
\draw (copierD) to[out=45,in=-90] ([xshiftu=-0.25]gra.south);
\draw (copierG) to[out=165,in=-90] (-1,3.5);
\draw (copierG) to[out=45,in=-90] (let);
\draw (int) to (copierI1);
\draw (copierI1) to[out=165,in=-90] ([xshiftu=0.25]gra.south);
\draw (copierI1) to[out=45,in=-90] (copierI2);
\draw (copierI2) to[out=45,in=-90] (sat);
\draw (copierI2) to[out=165,in=-90] (2,3.5);
\draw (sat) to (3.5,3.5);
\draw (gra) to (copierG);
\draw (let) to (0.5,3.5);
\end{tikzpicture}}}$
\end{center}
\caption{Joint distribution for the student network from
  Figure~\ref{fig:student} obtained as string diagram with additional
  copiers for non-final nodes.}
\label{fig:studentjoint}
\end{figure}

Another interesting observation is that disintegration forms a
\emph{modular} procedure. The formula~\eqref{diag:disintegration}
shows that disintegration yields a new decomposition of a given state:
such a decomposition being a state itself, disintegration may be
applied again. In fact, this repeated application is how a joint state
on multiple domains gets represented as a Bayesian network.  The
channel-based approach understands this process uniformly as a
step-by-step transformation of a given channel (a state) into another,
equivalent channel (a Bayesian network). Once again, string diagrams
are a useful formalism for visualising such correspondence. For
instance, the joint state associated with the Student network from
Figures~\ref{fig:student} and~\ref{fig:studentstring} can be expressed
as in Figure~\ref{fig:studentjoint}. Notice that the diagram in
Figure~\ref{fig:studentjoint} is just the one in
Figure~\ref{fig:studentstring} where each non-final node has been made
externally accessible via additional copiers
$\copier$. Figure~\ref{fig:studentjoint} has the type of a joint
state. Its value can then be calculated via state transformation, via
a composite channel that can be ``read off'' directly from the graph
in Figure~\ref{fig:studentjoint}, namely:
\begin{equation}
\label{eqn:studentjoint}
\begin{array}{rcl}
\textsf{joint}
& \coloneqq &
\big((\idmap\otimes\idmap\otimes c_{L}\otimes\idmap\otimes\idmap) 
   \klafter (\idmap\otimes\Delta\otimes\idmap\otimes c_{S}) 
   \klafter (\idmap\otimes c_{G}\otimes \Delta) 
   \klafter\; (\Delta\otimes\Delta)\big) 
\\
& & \qquad
   \gg \big(\omega_{D}\otimes\omega_{I}\big).
\end{array}
\end{equation}

\noindent The tool \EfProb~\cite{ChoJ17b} has been designed precisely
to evaluate such systematic expressions.

\medskip

Now that we have a formal description of the relationship between joint states and Bayesian networks, we turn to comparing Bayesian inference in these two settings. For reasons of
simplicity, we concentrate on the binary case. Suppose we have a joint
state $\sigma\in\Dst(X\times Y)$, now with \emph{evidence} on $X$.  In
the present setting this evidence can be an arbitrary predicate $p\in
[0,1]^{X}$ and not only a point predicate $\indic{\{x\}}$, as usual.
We like to find out the distribution on $Y$, given the evidence $p$.
Informally, this may be written as $\Prob(Y\mid p)$. More precisely,
it is the second marginal of the state obtained by updating with
the weakened version $\fW(p) = p\otimes\one$, as in:
\[ \sM\big(\sigma\big|_{\subfW(p)}\big). \]

\noindent Now suppose we have factorised the joint state $\sigma$ as a
(mini) network~\eqref{diag:disintegration} via the extracted state
$c\colon \kto{X}{Y}$. We can also perfom causal reasoning ---
\textit{i.e.}~forward inference --- and obtain the state:
\[ c \gg \omega|_{p}
\mbox{\qquad where \qquad}
\begin{array}{rcl}
\omega 
& = &
\fM(\sigma).
\end{array} \]

\noindent The \emph{Bayesian inference theorem} says that these
outcomes are the same, not only for forward reasoning, but also for
backward reasoning.

\begin{theorem}
\label{thm:inference}
Let $\sigma\in\Dst(X\times Y)$ be a joint state with extracted channel
$c\colon \kto{X}{Y}$ as in~\eqref{diag:disintegration}. For predicates
$p\in [0,1]^{X}$ and $q\in [0,1]^{Y}$ one has:
\[ \begin{array}{rclcrcl}
\sM\big(\sigma\big|_{\subfW(p)}\big)
& = &
c \gg \fM(\sigma)\big|_{p}
& \mbox{\qquad and \qquad} &
\fM\big(\sigma\big|_{\subsW(q)}\big)
& = &
\fM(\sigma)\big|_{c \ll q}.
\end{array} \]
\end{theorem}

Before giving a proof, we comment on the significance of the
statement. Inference with joint states, as on the left-hand-side of
the equations in Theorem~\ref{thm:inference}, involves weakening
$\weaksign$ of evidence in one coordinate and marginalisation
$\margsign$ in another coordinate. It uses the entwinedness of the
joint stage $\sigma$, so that one coordinate can influence the other,
see~\cite{JacobsZ17} where this is called \emph{crossover}
influence. Therefor we like this call this form of inference via joint
states \emph{crossover inference}.

In contrast, inference on the right-hand-side of the equations in
Theorem~\ref{thm:inference} essentially uses state and predicate
transformation $\gg$ and $\ll$.  Therefor we refer to this form of
inference as \emph{transformer inference}. It consists of what we have
called backward and forward inference in the table at the end of
Subsection~\ref{subsec:transformation}.

Thus the Bayesian inference theorem states the equivalence of
crossover inference and transformer inference. Whereas crossover
inference works well with small samples (see the examples below), it
does not scale to larger networks, where transformations inference is
preferable. The equivalence is widely known at some implicit level,
but its formulation in this explicit form only arises within the
current channel-based perspective on Bayesian networks.

We now provide a proof of the theorem. A purely diagrammatic argument
is given in~\cite{ChoJ17a}.

\begin{myproof}[Proof of Theorem \ref{thm:inference}]
We confine ourselves to proving the first equation in concrete form,
using the definition of extracted channel
from~\eqref{eqn:disintegration}:
$$\begin{array}[b]{rcl}
\lefteqn{\big(c \gg \fM(\sigma)\big|_{p}\big)(y)}
\\
& = &
\displaystyle\sum_{x} c(x)(y) \cdot \fM(\sigma)\big|_{p}(x)
\\
& \smash{\stackrel{\eqref{eqn:disintegration}}{=}} &
\displaystyle\sum_{x} \frac{\sigma(x,y)}{\fM(\sigma)(x)} 
   \cdot \frac{\fM(\sigma)(x)\cdot p(x)}{\fM(\sigma)\models p}
\\
& \smash{\stackrel{\eqref{eqn:marginalisationweakening}}{=}} &
\displaystyle\sum_{x} \frac{\sigma(x,y)\cdot p(x)}{\sigma\models \fW(p)}
\\
& = &
\displaystyle\sum_{x} \frac{\sigma(x,y)\cdot \fW(p)(x,y)}
                          {\sigma \models \fW(p)}
   \quad \mbox{since } \fW(p)(x,y) = (p\otimes\one)(x,y) = p(x)
\\
& = &
\displaystyle\sum_{x} \sigma\big|_{\subfW(p)}(x,y)
\\
& = &
\sM\big(\sigma\big|_{\subfW(p)}\big)(y).
\end{array}\eqno{\QEDbox}$$
\end{myproof}

We conclude this section by giving two demonstrations of the
equivalence stated in Theorem~\ref{thm:inference}. First, we answer
once again the six questions about the student network in
Section~\ref{sec:bayesiannetwork}: whereas therein we applied
transformer inference, we now compute using crossover inference. We
shall write:
\[ \textsf{joint} \,\in\, \Dst\Big(D\times G\times L \times I\times S\Big) \]

\noindent for the joint state associated with the student network,
obtained in formula~\eqref{eqn:studentjoint}, following
Figure~\ref{fig:studentjoint}. We write $\margsign_{i}$ for the $i$-th
marginal, obtained by summing out all domains which are not in the
$i$-th position. For the sake of clarity we do not use the notation
$\weaksign$ for weakening, but use parallel product with the truth
predicate $\one$ instead. In agreement with
Theorem~\ref{thm:inference}, The outcomes are the same as in
Section~\ref{sec:bayesiannetwork}, but they have been computed
separately (in \EfProb).
\begin{enumerate}
\item \textbf{What are the a priori probabilities
  for the recommendation?}
\[ \begin{array}{rcl}
\margsign_{3}(\textsf{joint})
& = &
0.498\ket{l^0} + 0.502\ket{l^1}.
\end{array} \]

\item \textbf{What if we know that the student is
  not intelligent?} 
\[ \begin{array}{rcl}
\margsign_{3}\big(\textsf{joint}
   \big|_{\one\otimes\one\otimes\one\otimes\indic{\{i^0\}}\otimes\one}\big)
& = &
0.611\ket{l^0} + 0.389\ket{l^1}.
\end{array} \]

\item \textbf{What if we additionally know that the test is easy?}
\[ \begin{array}{rcl}
\margsign_{3}\big(\textsf{joint}
   \big|_{\indic{\{d^0\}}\otimes\one\otimes\one\otimes\indic{\{i^0\}}\otimes\one}\big)
& = &
0.487\ket{l^0} + 0.513\ket{l^1}.
\end{array} \]

\item \textbf{What is the intelligence given a C-grade ($g^3$)?} 
\[ \begin{array}{rcl}
\margsign_{4}\big(\textsf{joint}
   \big|_{\one\otimes\indic{\{g^3\}}\otimes\one\otimes\one\otimes\one}\big)
& = &
0.921\ket{i^0} + 0.0789\ket{i^1}.
\end{array} \]

\item \textbf{What is the intelligence given a weak recommendation?}  
\[ \begin{array}{rcl}
\margsign_{4}\big(\textsf{joint}
   \big|_{\one\otimes\one\otimes\indic{\{l^0\}}\otimes\one\otimes\one}\big)
& = &
0.86\ket{i^0} + 0.14\ket{i^1}.
\end{array} \]

\item \textbf{What is the intelligence given a C-grade but a high SAT score?}
\[ \begin{array}{rcl}
\margsign_{4}\big(\textsf{joint}
   \big|_{\one\otimes\indic{\{g^3\}}\otimes\one\otimes\one\otimes\indic{\{s^1\}}}\big)
& = &
0.422\ket{i^0} + 0.578\ket{i^1}.
\end{array} \]
\end{enumerate}


As a second demonstration of the theorem, we briefly return to the controversy around inference with
soft predicates in Example~\ref{ex:burglar}.

\begin{example}
\label{ex:burglarjoint}
We first re-arrange the Bayesian network from Example~\ref{ex:burglar}
in string diagrammatic form so that we can compute the joint state
$\omega\in\Dst(B\times A\times E\times R)$ as:
\[ \omega
\;\;
\coloneqq
\;\;
\vcenter{\hbox{%
\begin{tikzpicture}[font=\small]
\node[state] (B) at (-1,0) {$\omega_B$};
\node[copier] (copierB) at (-1,0.2) {};
\node[state] (E) at (1,0) {$\omega_E$};
\node[copier] (copierE1) at (1,0.2) {};
\node[copier] (copierE2) at (1.5,0.5) {};
\node[arrow box] (A) at (0,1) {$\;c_{A}\;$};
\node[arrow box] (R) at (2.0,1) {$\;c_{R}\;$};
\draw (B) to (copierB);
\draw (copierB) to[out=165,in=-90] (-1.5,1.5);
\draw (copierB) to[out=45,in=-90] ([xshiftu=-0.2]A.south);
\draw (E) to (copierE1);
\draw (copierE1) to[out=165,in=-90] ([xshiftu=0.2]A.south);
\draw (copierE1) to[out=45,in=-90] (copierE2);
\draw (copierE2) to[out=45,in=-90] (R);
\draw (copierE2) to[out=165,in=-90] (1,1.5);
\draw (A) to (0,1.5);
\draw (R) to (2.0,1.5);
\end{tikzpicture}}}
\hspace*{3em}\mbox{\textit{i.e.}}\hspace*{2.5em}
{\begin{array}{rcl}
\omega
& = &
\big((\idmap \otimes \idmap \otimes c_{R} \otimes \idmap)
\\
& & \qquad \klafter\; (\idmap\otimes c_{A} \otimes \Delta)
\\
& & \qquad \klafter\; (\Delta\otimes\Delta)\big)
   \gg (\omega_{B}\otimes\omega_{E}).
\end{array}}
\]

\noindent Recall that we have soft/fuzzy evidence $q(a) = 0.7,
q(\no{a}) = 0.3$ on $A$. Given this evidence, we want to know the
burglar probability. Using crossover inference it is computed as:
\[ \begin{array}{rcl}
\margsign_{1}\big(\omega\big|_{\one\otimes q\otimes \one\otimes \one}\big)
& = &
0.0229\ket{b} + 0.9771\ket{\no{b}}.
\end{array} \]

\noindent We obtain the same outcome as via transformer infererence
in~\eqref{eqn:softpullback}. Of course, the Bayesian Inference
Theorem~\ref{thm:inference} tells that the outcomes should coincide in
general. This additional computation just provides further support for
the appropriateness of doing inference via forward and backward
transformations along channels.
\end{example}

\section{Conclusions}\label{sec:conclusions}

This chapter provides an introduction to an emerging area of
channel-based probability theory. It uses standard compositional
techniques from programming semantics in the area of Bayesian
inference, giving a conceptual connection between forward and backward
inference (or: causal and evidential reasoning) on the one hand, and
crossover influence on the other.

Promising research directions within this framework include the
development of channel-based algorithms for Bayesian
reasoning~\cite{Jacobs18b}. Moreover, the abstract perspective offered
by the channel approach may apply to probabilistic graphical models
other than Bayesian networks, including models for machine learning
such as neural networks (see~\cite{JacobsS18} for first steps). Paired
with the mathematical language of string diagrams, this framework may
eventually offer a unifying compositional perspective on the many
different pictorial notations for probabilistic reasoning.

\medskip

\noindent\textbf{Acknowledgements} Fabio Zanasi acknowledges support
from EPSRRC grant nr.~EP/R020604/1. Bart Jacobs' research has received
funding from the European Research Council under the European Union's
Seventh Framework Programme (FP7/2007-2013) / ERC grant agreement
nr.~320571


\begin{thebibliography}{10}

\bibitem{Barber12}
D.~Barber.
\newblock {\em Bayesian Reasoning and Machine Learning}.
\newblock Cambridge Univ. Press, 2012.
\newblock publicly available via
  \url{http://web4.cs.ucl.ac.uk/staff/D.Barber/pmwiki/pmwiki.php?n=Brml.HomePage}.

\bibitem{BartelsSV04}
F.~Bartels, A.~Sokolova, and E.~de~Vink.
\newblock A hierarchy of probabilistic system types.
\newblock {\em Theoretical Computer Science}, 327(1-2):3--22, 2004.

\bibitem{ChoJ17a}
K.~Cho and B.~Jacobs.
\newblock Disintegration and {Bayesian} inversion, both abstractly and
  concretely.
\newblock See \url{arxiv.org/abs/1709.00322}, 2017.

\bibitem{ChoJ17b}
K.~Cho and B.~Jacobs.
\newblock The {EfProb} library for probabilistic calculations.
\newblock In F.~Bonchi and B.~K{\"o}nig, editors, {\em Conference on Algebra
  and Coalgebra in Computer Science (CALCO 2017)}, volume~72 of {\em LIPIcs}.
  Schloss Dagstuhl, 2017.

\bibitem{ChoJWW15}
K.~Cho, B.~Jacobs, A.~Westerbaan, and B.~Westerbaan.
\newblock An introduction to effectus theory.
\newblock see \url{arxiv.org/abs/1512.05813}, 2015.

\bibitem{CulbertsonS14}
J.~Culbertson and K.~Sturtz.
\newblock A categorical foundation for bayesian probability.
\newblock {\em Appl. Categorical Struct.}, 22(4):647--662, 2014.

\bibitem{DijkstraS90}
E.~Dijkstra and C.~Scholten.
\newblock {\em Predicate Calculus and Program Semantics}.
\newblock Springer, Berlin, 1990.

\bibitem{Dijkstra:1997}
E.~W. Dijkstra.
\newblock {\em A Discipline of Programming}.
\newblock Prentice Hall PTR, Upper Saddle River, NJ, USA, 1st edition, 1997.

\bibitem{DuboisP90}
D.~Dubois and H.~Prade.
\newblock The logical view of conditioning and its application to possibility
  and evidence theories.
\newblock {\em Int. Journ. of Approximate Reasoning}, 4:23--46, 1990.

\bibitem{Fong12}
B.~Fong.
\newblock Causal theories: A categorical perspective on {Bayesian} networks.
\newblock Master's thesis, Univ.\ of Oxford, 2012.
\newblock see \url{arxiv.org/abs/1301.6201}.

\bibitem{Giry82}
M.~Giry.
\newblock A categorical approach to probability theory.
\newblock In B.~Banaschewski, editor, {\em Categorical Aspects of Topology and
  Analysis}, number 915 in Lect. Notes Math., pages 68--85. Springer, Berlin,
  1982.

\bibitem{Jacobs11c}
B.~Jacobs.
\newblock Probabilities, distribution monads, and convex categories.
\newblock {\em Theoretical Computer Science}, 412(28):3323--3336, 2011.

\bibitem{Jacobs13a}
B.~Jacobs.
\newblock Measurable spaces and their effect logic.
\newblock In {\em Logic in Computer Science}. IEEE, Computer Science Press,
  2013.

\bibitem{Jacobs17a}
B.~Jacobs.
\newblock From probability monads to commutative effectuses.
\newblock {\em Journ. of Logical and Algebraic Methods in Programming},
  94:200--237, 2017.

\bibitem{Jacobs17b}
B.~Jacobs.
\newblock A recipe for state and effect triangles.
\newblock {\em Logical Methods in Comp. Sci.}, 13(2), 2017.
\newblock See \url{https://lmcs.episciences.org/3660}.

\bibitem{Jacobs18b}
B.~Jacobs.
\newblock A channel-based exact inference algorithm for {Bayesian} networks.
\newblock See \url{arxiv.org/abs/1804.08032}, 2018.

\bibitem{JacobsS18}
B.~Jacobs and D.~Sprunger.
\newblock Neural nets via forward state transformation and backward loss
  transformation.
\newblock See \url{arxiv.org/abs/1803.09356}, 2018.

\bibitem{JacobsZ16}
B.~Jacobs and F.~Zanasi.
\newblock A predicate/state transformer semantics for {Bayesian} learning.
\newblock In L.~Birkedal, editor, {\em Math. Found. of Programming Semantics},
  number 325 in Elect. Notes in Theor. Comp. Sci., pages 185--200. Elsevier,
  Amsterdam, 2016.

\bibitem{JacobsZ17}
B.~Jacobs and F.~Zanasi.
\newblock A formal semantics of influence in bayesian reasoning.
\newblock In K.~Larsen, H.~Bodlaender, and J.-F. Raskin, editors, {\em Math.
  Found. of Computer Science}, volume~83 of {\em LIPIcs}. Schloss Dagstuhl,
  2017.

\bibitem{JonesP89}
C.~Jones and G.~Plotkin.
\newblock A probabilistic powerdomain of evaluations.
\newblock In {\em Logic in Computer Science}, pages 186--195. IEEE, Computer
  Science Press, 1989.

\bibitem{JungT98}
A.~Jung and R.~Tix.
\newblock The troublesome probabilistic powerdomain.
\newblock In A.~Edalat, A.~Jung, K.~Keimel, and M.~Kwiatkowska, editors, {\em
  Comprox III, Third Workshop on Computation and Approximation}, number~13 in
  Elect. Notes in Theor. Comp. Sci., pages 70--91. Elsevier, Amsterdam, 1998.

\bibitem{Keimel08}
K.~Keimel.
\newblock The monad of probability measures over compact ordered spaces and its
  {E}ilenberg-{M}oore algebras.
\newblock {\em Topology and its Applications}, 156:227--239, 2008.

\bibitem{KeimelP09}
K.~Keimel and G.~Plotkin.
\newblock Predicate transformers for extended probability and non-determinism.
\newblock {\em Math. Struct. in Comp. Sci.}, 19(3):501--539, 2009.

\bibitem{KollerF09}
D.~Koller and N.~Friedman.
\newblock {\em Probabilistic Graphical Models. Principles and Techniques}.
\newblock {MIT} Press, Cambridge, MA, 2009.

\bibitem{Kozen81}
D.~Kozen.
\newblock Semantics of probabilistic programs.
\newblock {\em Journ. Comp. Syst. Sci}, 22(3):328--350, 1981.

\bibitem{Kozen85}
D.~Kozen.
\newblock A probabilistic {PDL}.
\newblock {\em Journ. Comp. Syst. Sci}, 30(2):162--178, 1985.

\bibitem{Mermin07}
N.D. Mermin.
\newblock {\em Quantum Computer Science: An Introduction}.
\newblock Cambridge Univ. Press, 2007.

\bibitem{Mislove12}
M.~Mislove.
\newblock Probabilistic monads, domains and classical information.
\newblock In E.~Kashefi, J.~Krivine, and F.~van Raamsdonk, editors, {\em
  Developments of Computational Methods (DCM 2011)}, number~88 in Elect. Proc.
  in Theor. Comp. Sci., pages 87--100, 2012.

\bibitem{BenMradDPLA15}
A.~Ben Mrad, V.~Delcroix, S.~Piechowiak, P.~Leicester, and M.~Abid.
\newblock An explication of uncertain evidence in {Bayesian} networks:
  likelihood evidence and probabilistic evidence.
\newblock {\em Applied Intelligence}, 23(4):802--824, 2015.

\bibitem{Panangaden09}
P.~Panangaden.
\newblock {\em Labelled {Markov} Processes}.
\newblock Imperial College Press, London, 2009.

\bibitem{RusselN03}
S.~Russell and P.~Norvig.
\newblock {\em Artificial Intelligence. A Modern Approach}.
\newblock Prentice Hall, 2003.

\bibitem{ScibiorGG15}
A.~\'{S}cibior, Z.~Ghahramani, and A.~Gordon.
\newblock Practical probabilistic programming with monads.
\newblock In {\em Proc. 2015 ACM SIGPLAN Symp. on {Haskell}}, pages 165--176.
  ACM, 2015.

\bibitem{ScibiorKVSYCOMHG18}
A.~\'{S}cibior, O.~Kammar, M.~V\'{a}k\'{a}r, S.~Staton, H.~Yang, Y.~Cai,
  K.~Ostermann, S.~Moss, C.~Heunen, and Z.~Ghahramani.
\newblock Denotational validation of higher-order {Bayesian} inference.
\newblock In {\em Princ. of Programming Languages}, pages 60:1--60:29. ACM
  Press, 2018.

\bibitem{Selinger2009}
P.~Selinger.
\newblock A survey of graphical languages for monoidal categories.
\newblock {\em Springer Lecture Notes in Physics}, 13(813):289--355, 2011.

\bibitem{Sokolova11}
A.~Sokolova.
\newblock Probabilistic systems coalgebraically: A survey.
\newblock {\em Theoretical Computer Science}, 412(38):5095--5110, 2011.

\bibitem{StatonYHKW16}
S.~Staton, H.~Yang, C.~Heunen, O.~Kammar, and F.~Wood.
\newblock Semantics for probabilistic programming: higher-order functions,
  continuous distributions, and soft constraints.
\newblock In {\em Logic in Computer Science}. IEEE, Computer Science Press,
  2016.

\bibitem{TixKP05}
R.~Tix, K.~Keimel, and G.~Plotkin.
\newblock {\em Semantic Domains for Combining Probability and Non-Determinism}.
\newblock Number 129 in Elect. Notes in Theor. Comp. Sci. Elsevier, Amsterdam,
  2005.

\bibitem{ValtortaKV02}
M.~Valtorta, Y.-G. Kim, and J.~Vomlel.
\newblock Soft evidential update for probabilistic multiagent systems.
\newblock {\em Int. Journ. of Approximate Reasoning}, 29(1):71--106, 2002.

\bibitem{VaraccaW06}
D.~Varacca and G.~Winskel.
\newblock Distributing probability over non-determinism.
\newblock {\em Math. Struct. in Comp. Sci.}, 16:87--113, 2006.

\bibitem{VinkR99}
E.~de Vink and J.~Rutten.
\newblock Bisimulation for probabilistic transition systems: a coalgebraic
  approach.
\newblock {\em Theoretical Computer Science}, 221:271--293, 1999.

\end{thebibliography}

\end{document}